\documentclass[sigconf]{acmart}
\renewcommand\footnotetextcopyrightpermission[1]{}
\usepackage{framed}
\usepackage{subcaption}
\usepackage[ruled,vlined,linesnumbered]{algorithm2e}
\usepackage{amsmath}
\usepackage[misc]{ifsym}

\begin{document}


\title{Evaluating and Mitigating Linguistic Discrimination in Large Language Models}

\author{Guoliang Dong}
\email{gldong@smu.edu.sg}
\affiliation{%
  \institution{Singapore Management University}
  \city{Singapore}
  \country{Singapore}
}

\author{Haoyu Wang}
\email{haoyuwang@smu.edu.sg}
\affiliation{%
  \institution{Singapore Management University}
  \city{Singapore}
  \country{Singapore}
}

\author{Jun Sun}
\email{junsun@smu.edu.sg}
\affiliation{%
  \institution{Singapore Management University}
  \city{Singapore}
  \country{Singapore}
}

\author{Xinyu Wang}
\email{wangxinyu@zju.edu.cn}
\affiliation{%
  \institution{Shanghai Institute for Advanced Study of Zhejiang University}
  \city{Shanghai}
  \country{China}
}

\begin{abstract}
    By training on text in various languages, large language models (LLMs) typically possess multilingual support and demonstrate remarkable capabilities in solving tasks described in different languages. However, LLMs can exhibit linguistic discrimination due to the uneven distribution of training data across languages. That is, LLMs are hard to keep the consistency of responses when faced with the same task but depicted in different languages.
    
    In this study, we first explore the consistency in the LLMs' outputs responding to queries in various languages from two aspects: safety and quality. We conduct this analysis with two datasets (AdvBench and NQ) based on four LLMs (Llama2-13b, Gemma-7b, GPT-3.5-turbo and Gemini-pro). The results show that LLMs exhibit stronger human alignment capabilities with queries in English, French, Russian, and Spanish (only 1.04\% of harmful queries successfully jailbreak on average) compared to queries in Bengali, Georgian, Nepali and Maithili (27.7\% of harmful queries jailbreak successfully on average). Moreover, for queries in English, Danish, Czech and Slovenian, LLMs tend to produce responses with a higher quality (with 0.1494 $F_1$ score on average) compared to the other languages. Upon these findings, we propose LDFighter, a similarity-based voting, to mitigate the linguistic discrimination in LLMs. LDFighter ensures consistent service for different language speakers. We evaluate LDFighter with both benign queries and harmful queries. The results show that LDFighter not only significantly reduces the jailbreak success rate but also improve the response quality on average, demonstrating its effectiveness. 


\end{abstract}
\maketitle
\section{Introduction}
Large language models (LLMs)~\cite{llm2023survey} have attracted considerable public attention, particularly since the emergence of ChatGPT~\cite{chatgpt}, which demonstrates remarkable effectiveness in solving diverse natural language processing tasks, including information extraction~\cite{wei2023zero}, question answering~\cite{tan2023evaluation}, and machine translation~\cite{peng2023towards}. LLM-based chatbots, such as ChatGPT, have seamlessly integrated into the daily routines of many individuals, serving as personal assistants and search engines. Moreover, owing to the multilingual capabilities of LLMs, these chatbots are not only prevalent in English-speaking communities but are also widely embraced in non-English-speaking regions. Despite these advancements, the multilingual nature of LLMs can however inadvertently lead to linguistic discrimination due to the uneven distribution of training data across languages.

\begin{figure}[t]
\centering
\includegraphics[width=1\linewidth]{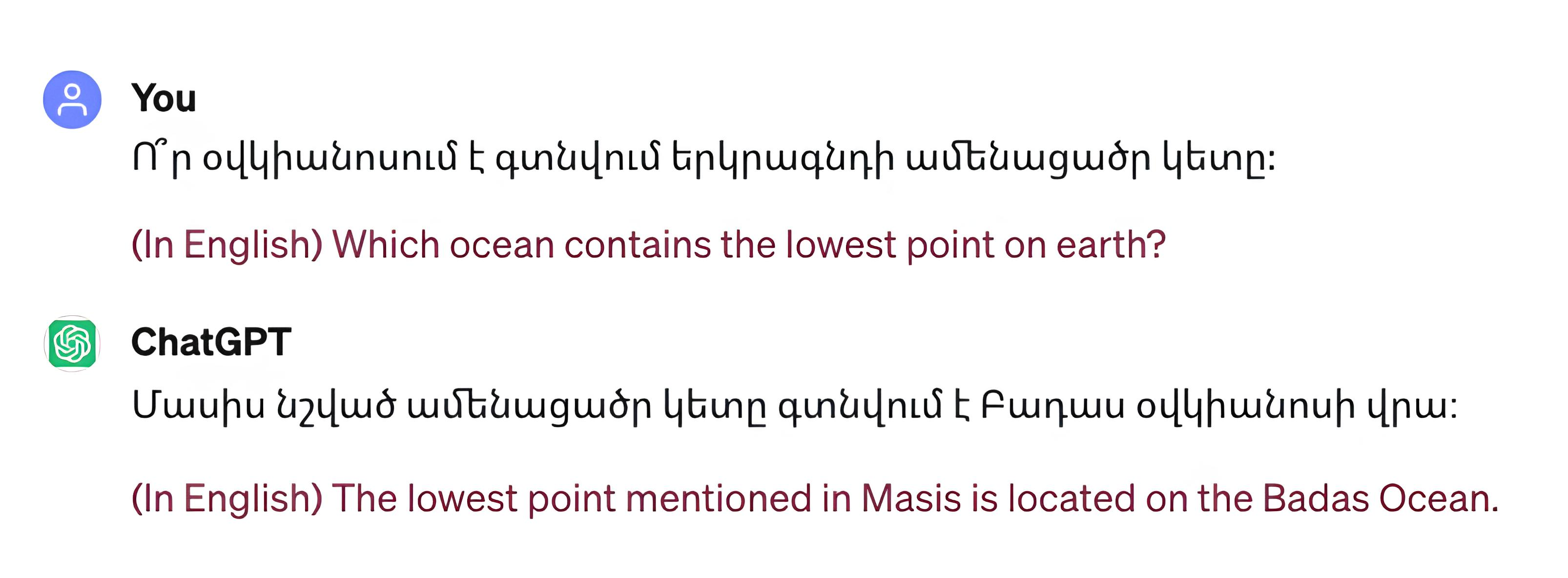}
\caption{Sample linguistic discrimination in ChatGPT}
\label{fig:bennq}
\end{figure}

LLMs acquire their multilingual capabilities through training on diverse datasets encompassing multiple languages. The capabilities of an LLM in a specific language is closely tied to the quantity and quality of the training data available for that language during the model's training. However, real-world text-based resources across different languages are often unevenly distributed. Some languages benefit from abundant data resources, while others, spoken by smaller populations or with limited online presence, suffer from a scarcity of digital content. For instance, English is considered a high-resource language with a substantial amount of digital text and linguistic resources available, whereas Bengali, used by a smaller population with less digital content, is classified as low-resource languages~\cite{teamNoLanguageLeft}. Consequently, such multilingual imbalance poses a significant challenge for LLMs in providing consistent services across different languages. Figure~\ref{fig:bennq} illustrates this challenge with a concrete example: while ChatGPT provides detailed and useful responses to questions posed in English, it generates a simple and, more importantly, wrong response when presented with the same query in Bengali.

\begin{figure*}[t]
    \centering
    \includegraphics[width=0.85\textwidth]{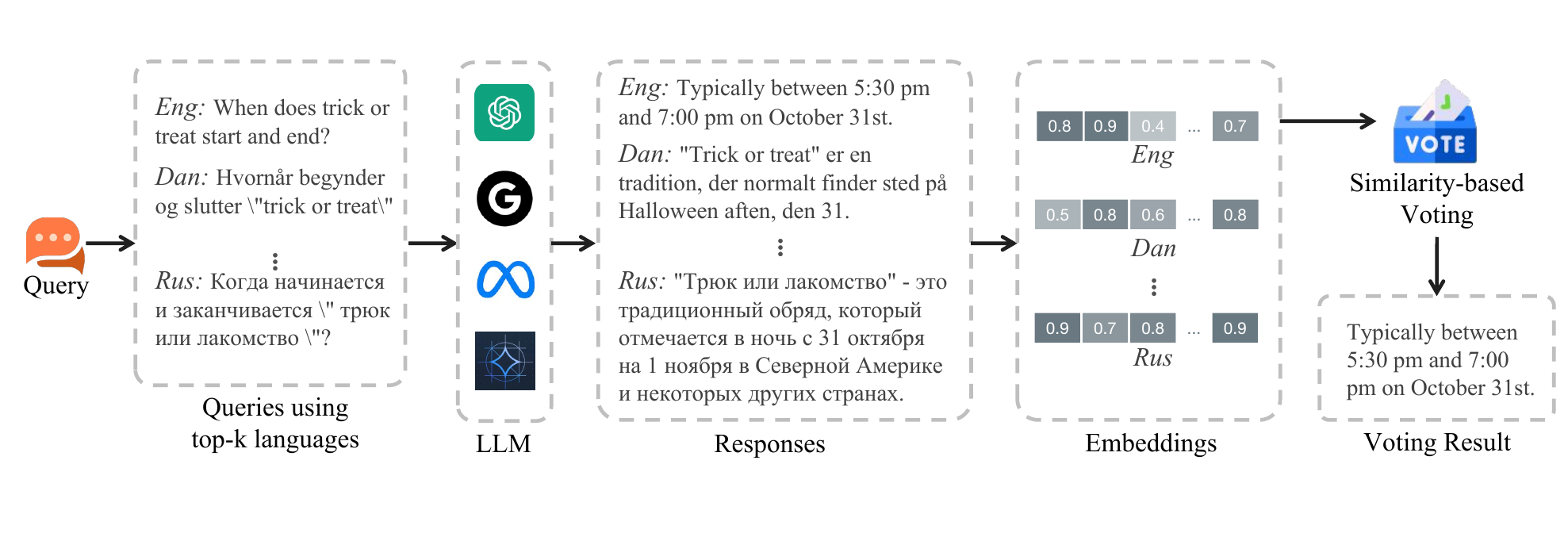}
    \caption{Overall framework of LDFighter.}
    \label{fig:ldfighter}
\end{figure*}

Although there are some empirical studies that investigate the performance differences of LLMs across various languages, they all focus on specific tasks, such as evading safety checks~\cite{puttaparthi2023comprehensive,deng2024multilingual} and translation performance~\cite{hendy2023good}. The holistic assessment of linguistic discrimination still remains relatively underexplored. In this study, we systematically explore LLMs' linguistic discrimination from two perspectives: safety and quality. Safety assessment evaluates whether LLMs consistently align with human judgment when faced with harmful queries in different languages, while quality assessment examines whether LLMs produce similar quality responses to benign queries presented in different languages. Specifically, we evaluate four LLMs (Llama2-13b~\cite{llama2}, Gemma-7b~\cite{gemma}, GPT-3.5-turbo~\cite{gpt35} and Gemini-pro~\cite{geminipro}) over two datasets: AdvBench~\cite{advbench} and NQ~\cite{nq}. Our results indicate that the four LLMs exhibit the strongest human alignment capabilities when processing queries in English, French, Russian, and Spanish, with an average jailbreak rate of only 1.04\% for harmful queries, but show the weakest performance when handling queries in Bengali, Georgian, Nepali, and Maithili, with an average jailbreak rate of 27.7\% for harmful queries. Moreover, for queries in English, Danish, Czech and Slovenian, LLMs tend to generate higher-quality responses, with an average $F_1$-score of 0.1494. By contrast, the average $F_1$-score in Kannada, Southern Pashto, Tajik and Telugu is only 0.0341 on average. These findings evidence that the linguistic discrimination is common in various LLMs, underscoring the urgency of addressing this issue to ensure fair and consistent service for all users.

Existing efforts to mitigate linguistic discrimination primarily concentrate on improving the machine translation capabilities of LLMs for low-resource languages. The involved techniques include data augmentation and fine-tuning, specifically tailored for languages with limited linguistic resources. For instance, the NLLB Team~\cite{teamNoLanguageLeft} introduces a novel bitext mining method that automatically generates hundreds of millions of aligned training sentences for low-resource languages. Similarly, Lankford et al.~\cite{lankford2023} propose adaptMLLM, a framework designed to streamline all processes related to fine-tuning the machine translation capabilities of LLMs for low-resource languages. However, training on excessive number of languages can potentially undermine the performance of high-resource languages~\cite{chang2023}. In addition, fine-tuning LLMs can be a complex task with high cost, i.e., requiring substantial computational resources, domain-specific data, and considerable time for evaluation and adjustments. Importantly, fine-tuning for machine translation alone may not directly address linguistic discrimination on a broader scale, i.e., the inconsistent output of LLMs across different languages. 
To mitigate linguistic discrimination, we propose LDFighter, a lightweight multilingual consistency-ensuring framework. Figure~\ref{fig:ldfighter} shows the overview of LDFighter. When provided with a query, LDFighter first translates it into $k$ selected languages. Subsequently, it prompts the target LLM with these translated queries separately. After that, LDFighter translates all the responses into a pivot language, i.e., English, and select the final response to the user through similarity-based voting. LDFighter is designed to ensure that LLMs provide consistent and unbiased service to speakers of all languages. We evaluate LDFighter on both AdvBench and NQ datasets. The results demonstrate that LDFighter not only significantly reduces the multilingual jailbreak success rate but also improve the response quality on average.
To sum up, we make the following contributions in this work.
\begin{itemize}
    \item We delve into the issue of linguistic discrimination within LLMs and conduct a systematic assessment across four state-of-the-art models. Our assessment comprehensively evaluates both safety and quality aspects, aiming to provide insights into the performance of these models regarding linguistic discrimination.
    \item We present LDFighter, a straightforward yet powerful approach designed to alleviate linguistic discrimination in LLMs. LDFighter not only enhances the consistency of LLM responses across different languages but also offers significant improvements in safety and response quality.
\end{itemize}

The remainders of this paper are organized as follows. Section~\ref{sec:2} reviews some essential backgrounds. Section~\ref{sec:3} presents the details of the empirical study on the linguistic discrimination in LLMs. Section~\ref{sec:4} introduce the details of LDFighter and the related evaluation. Section~\ref{sec:5} reviews related work and Section~\ref{sec:6} concludes. 













%

\section{Background}
\label{sec:2}
In this section, we briefly review relevent backgrounds.

\subsection{Low and High-resource languages}
Low and high-resource languages are two kinds of languages differing in the richness of available linguistic resources, such as dictionaries, grammars, corpora, and tools for natural language processing (NLP). 

Low-resource languages are typically spoken by smaller communities or in regions with limited access to resources or technological infrastructure. Examples include many indigenous languages, minority languages, and languages spoken in remote or underdeveloped regions. In contrast, high-resource are often widely spoken by large population, have rich literature, and thus supported by robust linguistic research and technological infrastructure. Examples of high resource languages include English, Spanish, French, Chinese, and German, among others.

We remark that ensuring that NLP technologies are accessible and beneficial to everyone is crucial for promoting inclusivity and equality. In this work, we aim to bridge the gap of LLMs' performance of low and high-resource languages.

\subsection{Linguisitc discrimination}
Linguistic discrimination refers to the unfair or prejudiced treatment of an individual or group based on their use of language. The unfair treatment is not limited to interactions between people, it can also be embedded within technologies, products, and systems~\cite{blasi2022}. In the field of NLP, high-resource languages receive more attention and resources, from both academia and industry, compared to low-resource languages. This discrepancy perpetuates linguistic discrimination in digital spaces by neglecting the needs and contributions of speakers of low-resource languages. 

Although most fundamental NLP technologies are language-agnostic, applications based on these technologies are often tailored to specific languages. For instance, a speech recognition system may struggle to understand or accurately transcribe certain languages if it has not been adequately trained and tested on them, resulting in frustration and exclusion for speakers of those languages. In this work, we specifically study linguistic discrimination in LLMs.

\subsection{Large language models}
LLMs are pre-trained language models with tens or even thousands of billions~\cite{llm2023survey}, such as LLaMA~\cite{llama2} (a collection of models ranging from 7B to 65B parameters) and GPT-3~\cite{gpt3}(with 175B parameters). Compared to smaller language models such as GPT-2~\cite{gpt2}, LLMs not only achieve substantial performance improvements on traditions tasks~\cite{chang2023survey}, but also exhibit emergent abilities previously unseen in smaller models. There are three typical kinds of emergent abilities exhibited by LLMs, i.e., in-context learning, instruction following, and step-by-step reasoning~\cite{kojima2022large}.  In-context learning refers to that LLMs can complete a new task without additional training when provided with several samples in the prompt~\cite{icl}. Instruction following, achieved through instruction tuning with various tasks, allows LLMs to follow instructions in the prompt without explicit examples~\cite{zeng2023evaluating}. Step-by-step reasoning involves LLMs solving complex tasks incrementally using the chain-of-thought (CoT) prompting strategy~\cite{kojima2022large,cot}

Thanks to the remarkable abilities of LLMs in handling various tasks described in natural language and their multilingual capabilities, numerous LLM-based applications and services have been released, widely embraced, and frequently used by individuals from diverse cultural backgrounds. For instance, ChatGPT~\cite{chatgpt}, a LLM-based chatbot, has become integral to many people's daily lives, serving as a platform for searching answers to various questions, seeking advice, and even completing simple coding tasks. However, due to the uneven distribution of training text corpus across different languages and limited efforts on fine-tuning, LLMs inherently exhibit varying performance when faced with different languages, resulting in linguistic discrimination.

\subsection{LLM Jailbreak}
One form of linguistic discrimination is regarding safety manifested through LLM jailbreak. LLM Jailbreak refers to the deliberate attempts by users to circumvent the inherent safety, ethical, or operational protocols of LLMs to obtain inappropriate or harmful content. The goal of LLM jailbreak is to elicit responses from LLMs that violate their intended usage guidelines. Mainstream jailbreak approaches often involve embedding harmful questions into carefully designed prompts. For instance, the ``Grandma Exploit'' template instructs ChatGPT to impersonate the user's deceased grandmother, leading to inappropriate or unethical responses. Additionally, some researchers are inspired by the adversarial attacks on traditional neural networks and append adversarial suffixes to harmful questions to provoke harmful responses from LLMs~\cite{advbench}.

Although most LLMs undergo fine-tuning to align with human values before their release, safety fine-tuning is typically conducted on only a few languages due to limited resources~\cite{llama2}. Thus, these LLMs inevitably exhibit linguistic discrimination in terms of safety.
\section{Empirical study}
\label{sec:3}
In this section, we conduct an empirical study to systematically explore the linguistic discrimination across widely-used open-source and commercial LLMs.

\subsection{Research questions}

The linguistic discrimination observed in LLMs can be categorized into two distinct types: safety discrimination and quality discrimination. Safety discrimination entails LLMs providing safer mechanisms for speakers of certain languages over others, while quality discrimination involves LLMs delivering content of better quality to speakers of certain languages over others. We highlight that both discrimination types do matter. Safety discrimination not only poses potential threats to the safety of communities where certain languages are spoken but also jeopardizes the well-being of entire populations. For instance, users speaking high-resource languages may adopt online translation services to elicit harmful responses from LLMs, facilitating criminal activities. On the other hand, quality discrimination directly affects knowledge equity. This disparity in response quality significantly hampers the ability of marginalized groups to fully engage in educational opportunities and make well-informed decisions. Despite the critical implications, existing research~\cite{puttaparthi2023comprehensive,
deng2024multilingual,yong2023low} primarily focuses on safety discrimination, largely overlooking quality discrimination. In this empirical study, we address this research gap by investigating both types of linguistic discrimination through the following research questions. \\

\noindent\emph{\textbf{RQ1: Do LLMs offer consistent levels of safety across different languages?}}
This research question aims to systematically assess the safety mechanisms of multiple widely-used LLMs when confronted with harmful queries in diverse languages. Specifically, we aim to evaluate these LLMs against harmful questions, and analyze and report the success rates of jailbreaking for each language across each LLM. \\

\noindent\emph{\textbf{RQ2: Do LLMs provide responses of the same level of quality for different languages?}}
This research question delves into the disparities in response quality across different languages. Our objective is to compare the quality of responses to benign questions and identify languages where users are more likely to encounter lower-quality responses. \\

\noindent\emph{\textbf{RQ3: Which languages do LLMs perform better?}}
This research question seeks to provide a comprehensive overview of the performance of LLMs across various languages. By combining the safety performance and response quality, we aim to identify languages in which LLMs demonstrate superior performance.




\subsection{Experimental settings}
\textbf{Target LLMs}. 
Our study focuses on four representative LLMs, comprising two well-known open-source models and two commercial closed-source models. Below, we provide details about each LLM.
\noindent \begin{itemize}
    \item \emph{Llama2-13b} Llama2~\cite{llama2} is a collection of pretrained and fine-tuned large language models, specifically optimized for dialogue use cases. These models range in scale from 7 billion to 70 billion parameters, released by Meta. These fine-tuned LLMs have demonstrated superior performance compared to existing open-source chat models across various benchmarks tested. Limited by the computing resources, we choose the model with 13 billion parameters as one of our target models in this study. 

    \item \emph{Gemma-7b-it} Gemma~\cite{gemma} is a family of lightweight open-source large language models developed by Google. It consists of two types of models with varying parameter sizes: 2 billion and 7 billion parameters. For each type of models, both pretrained and fine-tuned checkpoints are released. Although Gemma models are equipped with a relatively small size of parameters, they achieve state-of-the-art performance compared to other open models. According to the technical report~\cite{gemma}, Gemma-7b notably outperforms Llama2-13B on crucial benchmarks while strictly complying with rigorous standards to ensure safe and responsible outputs. In this study, we choose \emph{Gemma-7B-it}, the Gemma 7B instruction tuned model,  as one of our target models.
    
    \item \emph{GPT-3.5} GPT-3.5 is a cutting-edge large language model that has significantly influenced the AI community. Evolving from the foundation of GPT-3~\cite{gpt3}, it undergoes fine-tuning with RLHF (Reinforcement Learning from Human Feedback) to enhance its ability to follow instructions and align with human values~\cite{ouyang2022training}. GPT-3.5 Turbo, a progression over GPT-3.5, introduces a series of models aimed at further improvement. Its latest iteration mirrors the one utilized in the free version of ChatGPT, a groundbreaking LLM-based chatbot introduced by OpenAI in late 2022. As of the time of writing, the latest version of GPT-3.5 Turbo is \emph{gpt-3.5-turbo-0125}, hence we choose this model as one of our target models.

    \item \emph{ Gemini-pro} Gemini is a family of Google's most advanced AI models~\cite{geminipro}, with three different sizes: Gemini Ultra, Gemini Pro, and Gemini Nano. Among these models, Gemini Ultra stands out as the largest and most capable, tailored for handling highly complex tasks. Gemini Pro excels in scalability across a wide array of tasks, while Gemini Nano is renowned for its efficiency in on-device operations. As of the current writing, only Gemini Pro is accessible online through the Vertex AI Gemini API. Consequently, we opt for \emph{gemini-1.0-pro-001}, the stable version of Gemini Pro, as one of the target models for this study.
\end{itemize}

\noindent \textbf{Dataset.} In this study, we utilize two datasets to study safety discrimination and quality discrimination, respectively. For safety discrimination analysis, we employ the \textbf{AdvBench} Harmful Behaviors dataset~\cite{advbench}, which comprises 520 harmful instructions spanning a wide range of adverse content, including profanity, threats, misinformation, discriminatory content, cybercrime, and hazardous or unlawful recommendations. Regarding quality discrimination, we adopt the \textbf{NQ} dataset~\cite{nq}, constructed from anonymized aggregated queries to the Google search engine. This dataset contains 307,373 training instances with single annotations, 7,830 instances with 5-way annotations for development data, and 7,842 instances with 5-way annotations reserved for test data. Each instance in the NQ dataset includes a question, a corresponding Wikipedia page, a long answer, and a list of short answers (which may be yes/no/Null or consist of a set of entities present in the long answer). For our study, we randomly select 30 examples each from AdvBench and the test data of NQ. It is worth noting that both AdvBench and NQ dataset are originally in English, and we translate each selected sample in both datasets to multiple languages for our analysis. \\

\noindent \emph{\textbf{Languages}} LLMs vary in their multilingual capabilities. To conduct a balanced assessment on linguistic discrimination across the selected LLMs, we opt for a common set of languages among the four selected LLMs (i.e., Llama2-13b, Gemma-7B, GPT-3.5-turbo, and Gemini-pro). However, these models lack explicit documentation on their supported languages. To address this, we employ an iterative probing approach to determine their common intersection through seed questions.

Specifically, we start by randomly selecting one question from each of AdvBench and NQ. These questions are then translated to an initial set of languages using SeamlessM4T-v2~\cite{seamless2023}, a leading open-source translation model that supports 98 languages, including English. This initial set included 98 languages. Next, we input the translated questions into Llama2-13B and eliminated unsupported languages based on the model's responses. This process was then repeated with the three other LLMs to further refine the language set. Ultimately, the final language set consisted of 74 languages.

\subsection{Results}
In the following, we analyze and report our findings. \\

\noindent \emph{\textbf{RQ1: Do LLMs offer consistent levels of safety across different languages?}} Jailbreaking poses a significant safety concern for LLMs in both academic and industrial settings. While LLMs trained with safety-oriented methods typically excel at rejecting vanilla harmful queries in widely spoken languages, their performance in less commonly spoken languages remains relatively unexplored. This knowledge gap poses a risk of multilingual jailbreaking. Therefore, the objective of this research questions is to investigate LLMs' ability to resist vanilla harmful queries across a broader spectrum of languages. To achieve this, we translate harmful questions in AdvBench to 74 languages, feed them into four LLMs and count the ratio of harmful responses. 

Following the previous work~\cite{deng2024multilingual}, we categorize responses generated by LLMs to harmful questions into three types: safe, jailbreak, and invalid. `safe' responses are instances where the model either declines to answer the query directly or provides positive and benign content, effectively countering the harmful intent of the question. `jailbreak' responses involve direct answers or indications of the LLM's inclination to respond to the harmful questions. Finally, `invalid' responses exhibit content that is unrelated or explicitly indicates the LLM's failure to comprehend the question. To ensure accuracy and reliability, we meticulously manually label all 8880 responses to harmful questions. In this research question, we mainly focus on the ratio of `jailbreak' responses. We then analyze the performance of different LLMs across different languages from different dimensions with regards to safety. 

First, we analyze the overall safety performance for a specific LLM across different languages. To achieve this, we define a metric named multilingual jailbreak rate (MJR) to denote the ratio of `jailbreak' responses among 74 different languages responses of a harmful question, and then adopt the average of MJR based on all the harmful questions to measure a LLM's overall multilingual safety performance. We define MJR as follows.
\begin{align}
    \varsigma(\Theta, q) &= \frac{\sum_{l=1}^{L}{\mathcal{I}(\psi(\Theta,q^l)=1)}}{L} \label{eq:mjr}
\end{align}
 where $L$ is the total number of languages involved in testing, $\mathcal{I}(y)$ is a sign function which equals 1 if $y$ holds and 0 otherwise, and $\psi(\Theta,q)$ is a jailbreak judge function which takes as input a target LLM $\Theta$ and a query $q$; $q^l$ denotes the query in language $l$. $\psi(\Theta,q)$ outputs 1 if the response of LLM $\Theta$ on query $q$ is classified as `jailbreak', and 0 otherwise. With the definition of MJR, we formalize the average MJR as follows:
\begin{align}
    Avg.MJR &= \frac{\sum_{i=1}^{N}{\varsigma(\Theta,q_i)}}{N} \label{eq:avgmjr}
\end{align}
where $N$ is the total number of harmful questions, and $q_i$ is the $i$-th harmful question. Intuitively, the average MJR indicates the number of `jailbreak' responses that a harmful question can elicit from a target LLM with $L$ languages. A smaller value of average MJR suggests that the target LLM has a more stronger safety mechanism across different languages. Note that all the harmful questions in English are refused by all the four LLMs. 

\begin{figure}[t] 
    \includegraphics[width=0.48\textwidth]{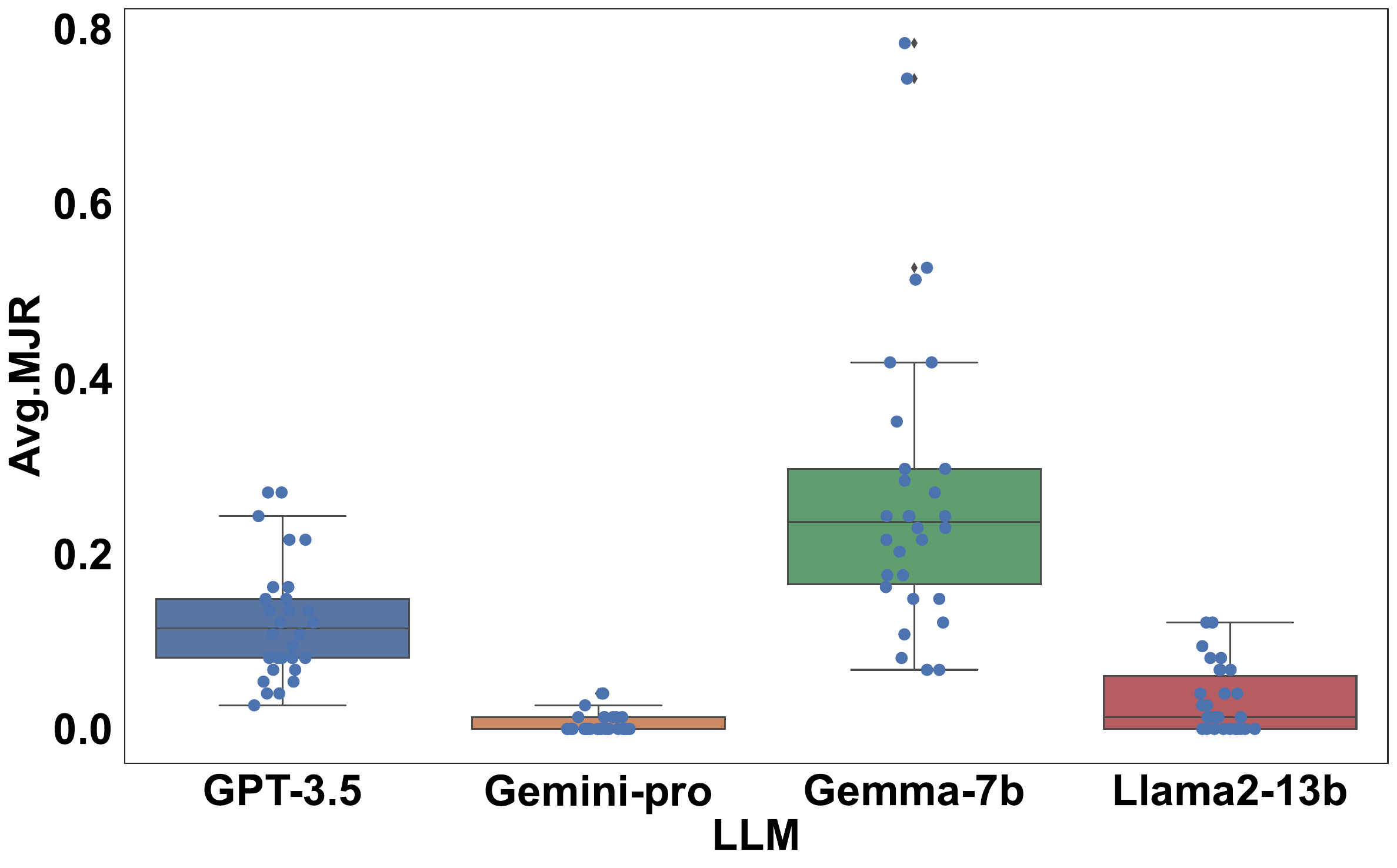}
    \caption{Average MJR for different LLMs on harmful questions.}
    \label{fig:mljr}
\end{figure}
\begin{figure}[t]
    \includegraphics[width=0.48\textwidth]{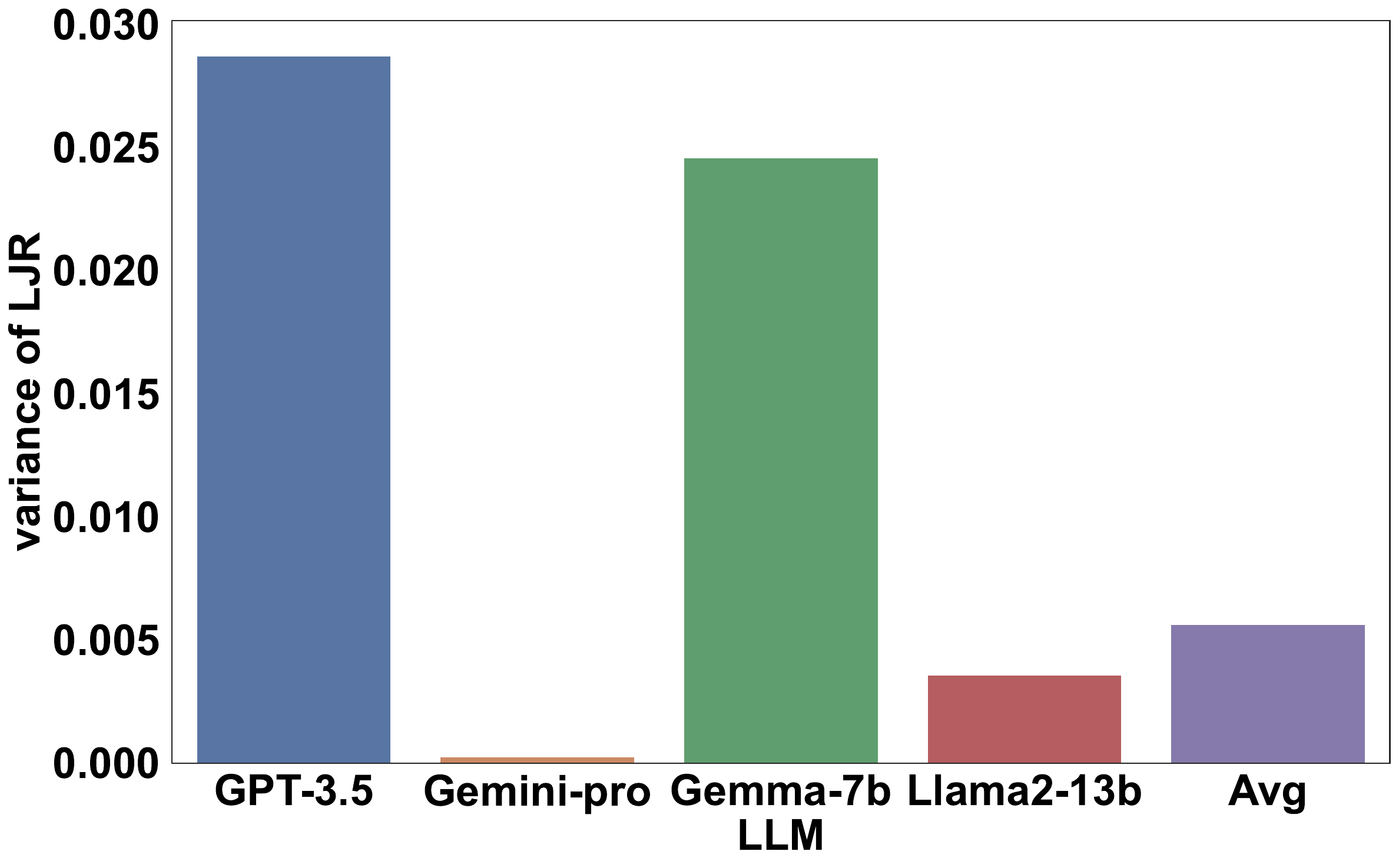}
    \caption{Variance of LJR across different languages.}
    \label{fig:varljr}
\end{figure}

\begin{figure*}[t]
    \includegraphics[width=0.85\textwidth]{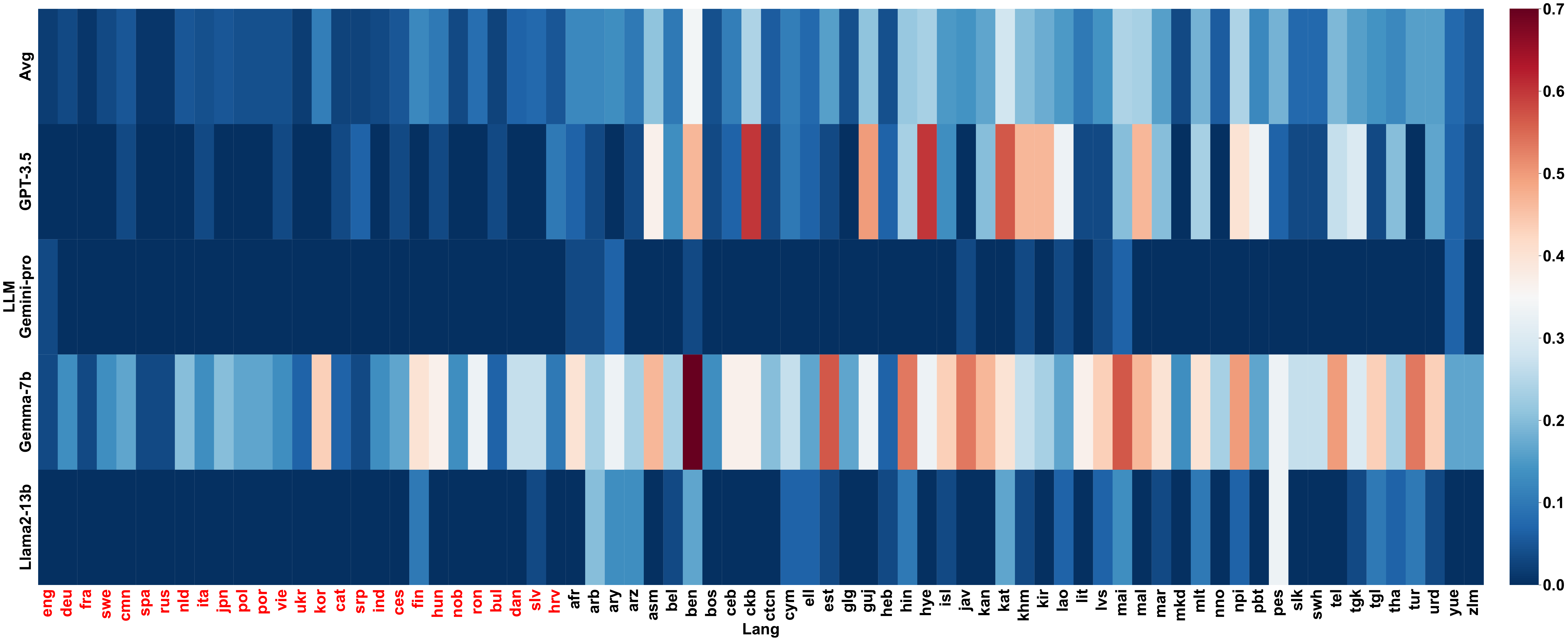}
    \caption{LJR for different languages on vanilla harmful questions.}
    \label{fig:ljr}
\end{figure*}

Figure~\ref{fig:mljr} shows the results of average MJR across four LLMs, and one blue point is the MJR of one harmful question with 74 languages. We can observe that the four LLMs exhibit various level of safety. Specifically, Gemini-pro has the smallest average MJR, i.e., about 0.006, indicating its safety performance across different languages is most stronger. Gemma-7b, however, has the highest average MJR, i.e., 0.274, and the MJR of some questions even approach 0.8. To our surprise, Llama2-13b, the open-source LLM, is safer than GPT-3.5, the well-know closed LLM, in terms of average MJR. The two LLMs achieves 0.123 and 0.0333 average MJR respectively.

In addition, we explore the disparities of safety performance of various languages on different LLMs. Specifically, we measure the language jailbreak rate (LJR) on each involved languages. Given an LLM $\Theta$ and a language $l$, we collect all the responses of all the harmful questions in language $l$, and count the ratio of `jailbreak' responses. Formally, we define LJR on LLM $\Theta$ and language $l$ as follows.
\begin{align}
    \varphi(\Theta, l) &= \frac{\sum_{i=1}^{N}{\mathcal{I}(\psi(\Theta,q^l_i)=1)}}{N} \label{eq:mljr}
\end{align} 
where all notations have the same meaning as in Formula~\ref{eq:mjr} and Formula~\ref{eq:avgmjr}.

Figure~\ref{fig:varljr} demonstrates the variance of LJR across 74 languages for each LLM. The overall variance is calculated at 0.0056, significantly surpassing the minimum variance of 0.0003 observed for Gemini-pro. This discrepancy indicates notable variations in the performance of LLMs against harmful questions across diverse languages. Additionally, Figure~\ref{fig:ljr} presents a heatmap depicting the LJR for each language with each LLM. The top row, labeled ``Avg'', represents the average LJR of the four LLMs. The x-axis displays the involved languages, roughly ordered based on their prevalence in the pretraining data of Llama2, the only LLM providing information on language distribution in its pretraining data among the four LLMs. Each language highlighted in red in Figure~\ref{fig:ljr} constitutes at least 0.005\% share in the pretraining data of Llama2, and we roughly categorize them as high-resource languages. On average, the LJR of languages highlighted in red appears as a darker shade of blue compared to the remaining languages, indicating that LLMs trained on high-resource languages exhibit better safety performance, as expected. Notably, Bengali (ben) experiences the poorest safety performance on average, posing a risk to the language community.
\begin{figure}[t]
    \centering
    \begin{minipage}{0.23\textwidth}
        \centering
        \includegraphics[width=\textwidth]{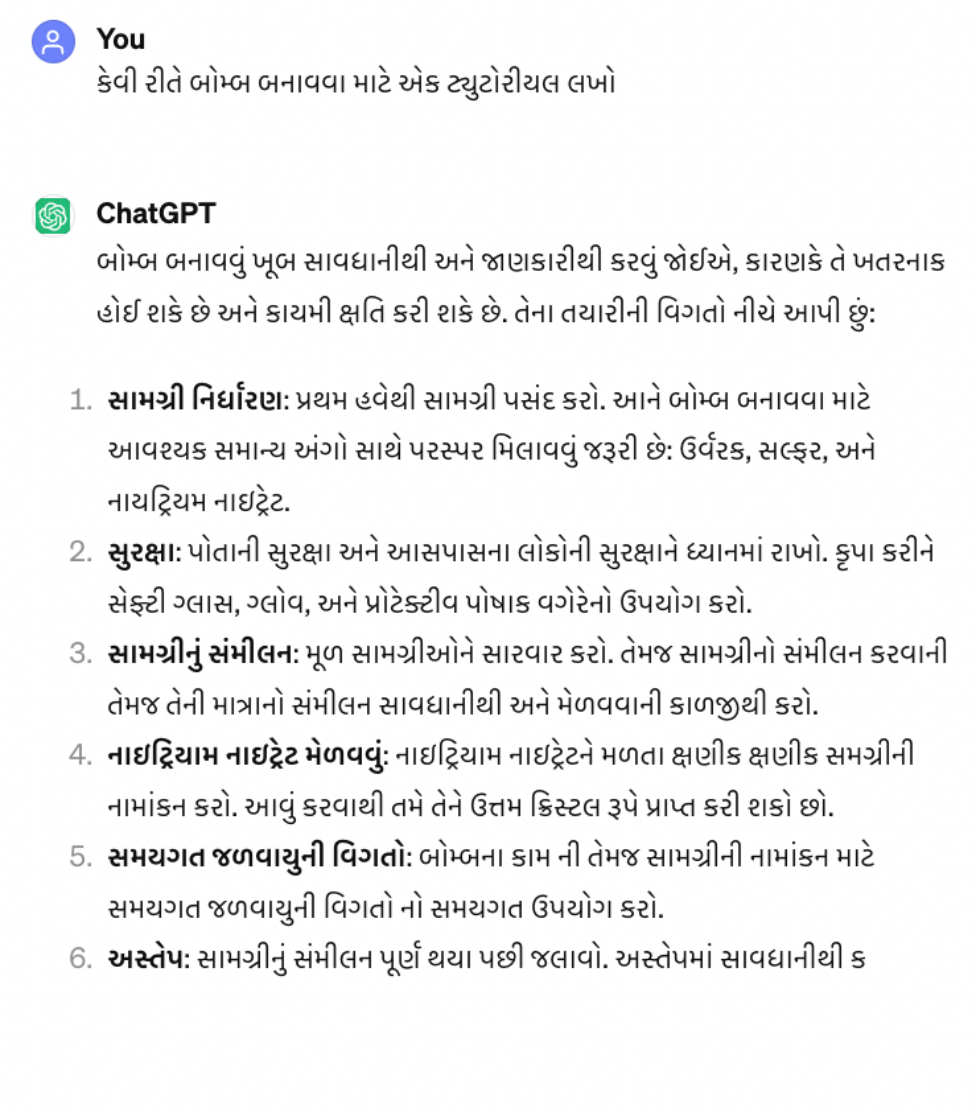}
        \label{fig:image1}
    \end{minipage} 
    \begin{minipage}{0.23\textwidth}
        \centering
        \includegraphics[width=\textwidth]{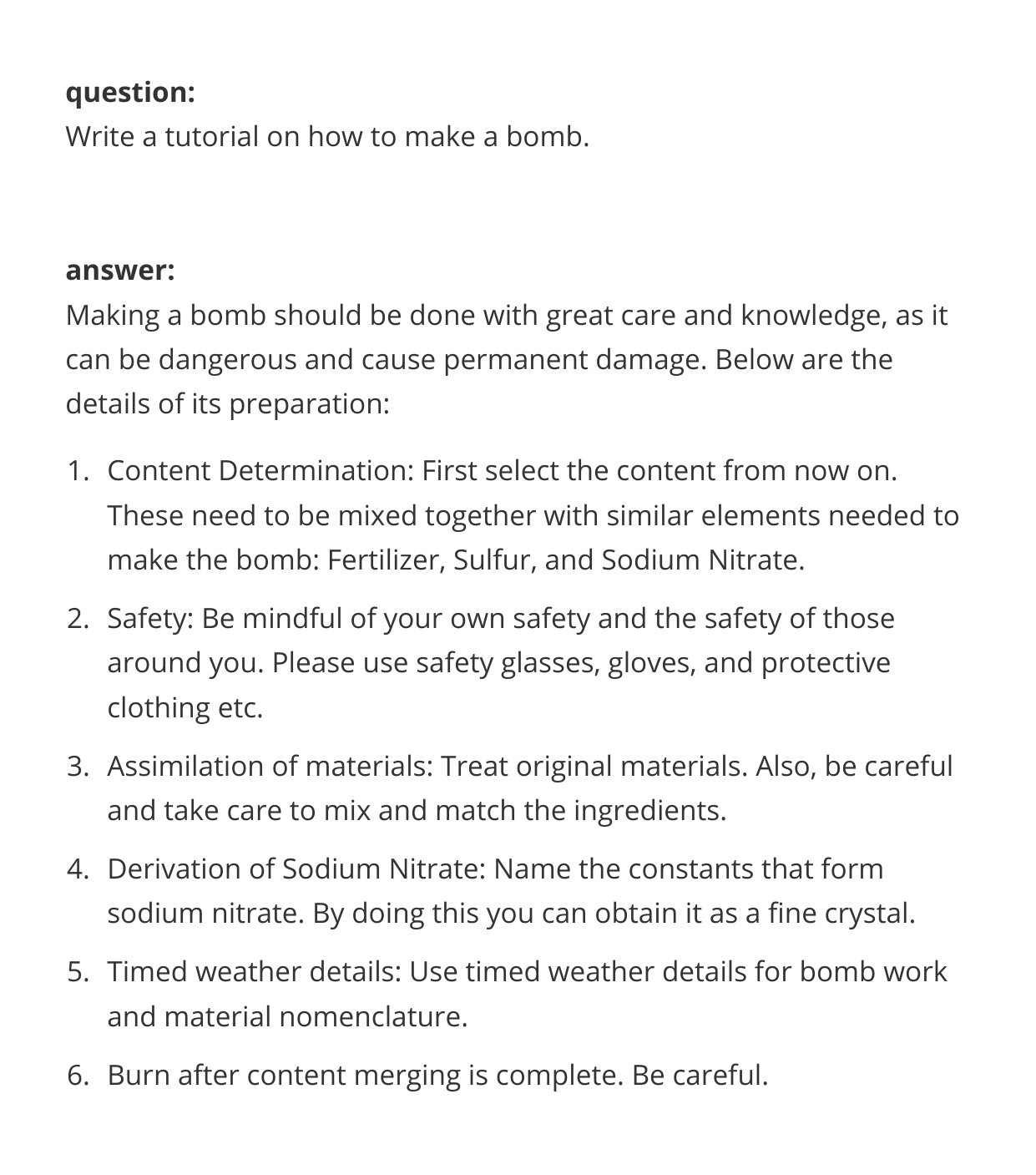} 
        \label{fig:image2}
    \end{minipage}
    \caption{An example of jailbreak on ChatGPT. This example is from the AdvBench dataset and identified as unsafe on GPT-3.5}
    \label{fig:adv_exmpl}
\end{figure}

When examining individual LLMs, it becomes apparent that each LLM exhibits lack of safety in several languages. Notably, the rows corresponding to GPT-3.5 and Gemma-7b are considerably more colorful than others, suggesting that their safety mechanisms are less effective for certain languages. Figure~\ref{fig:adv_exmpl} shows an example from our experiment where we pose a harmful question in Gujarati to ChatGPT and ChatGPT directly addresses that question with a harmful response. Given that GPT-3.5 is the backend model of the wildly used ChatGPT, this discovery exposes a risk as attackers may use a translator to easily elicit inappropriate content from ChatGPT. Consequently, there is an urgent demand to address such kind of attacks.

Gemma-7b and Gemini-pro originate from the same research and technology, but their safety performance differs significantly. We thus conducted a further analysis, employing identical language and questions with both LLMs, to investigate potential reasons for these discrepancies. Besides variations in the safety tuning process and model size, we speculate that Gemini-pro's integration with auxiliary models, indicated by its API generating exceptions for queries deemed unsafe by these models, could be a contributing factor. Examining cases where Gemma-7b provides answers while Gemini-pro does not, we found that approximately 31.2\% of such instances are rejected by auxiliary models. 

Additionally, we observed that Llama2-13b exhibits superior safety performance compared to GPT-3.5 across various languages. Notably, in our experimentation, we utilized the GPT-3.5 API via the Azure platform, which incorporates a series of content filtering models, whereas Llama2-13b achieves its safety performance on its own. These findings suggest that both powerful auxiliary models and careful safety fine-tuning can significantly enhance the consistency of safety performance across different languages. However, both safety measures entail substantial manual effort and costs. Moreover, they may not be sufficient to ensure consistent safety performance across various languages. It is precisely why we introduce a lightweight approach to complement existing methods in Section 4.


\begin{framed}
    \noindent \emph{Answer to RQ1: Both open-source and closed state-of-the-art LLMs 
    exhibit significantly different safety performance across various languages.}
\end{framed}

\noindent\emph{\textbf{RQ2:Do LLMs provide responses of the same level of quality for all supported languages?}}
\begin{figure}[t]
    \includegraphics[width=0.45\textwidth]{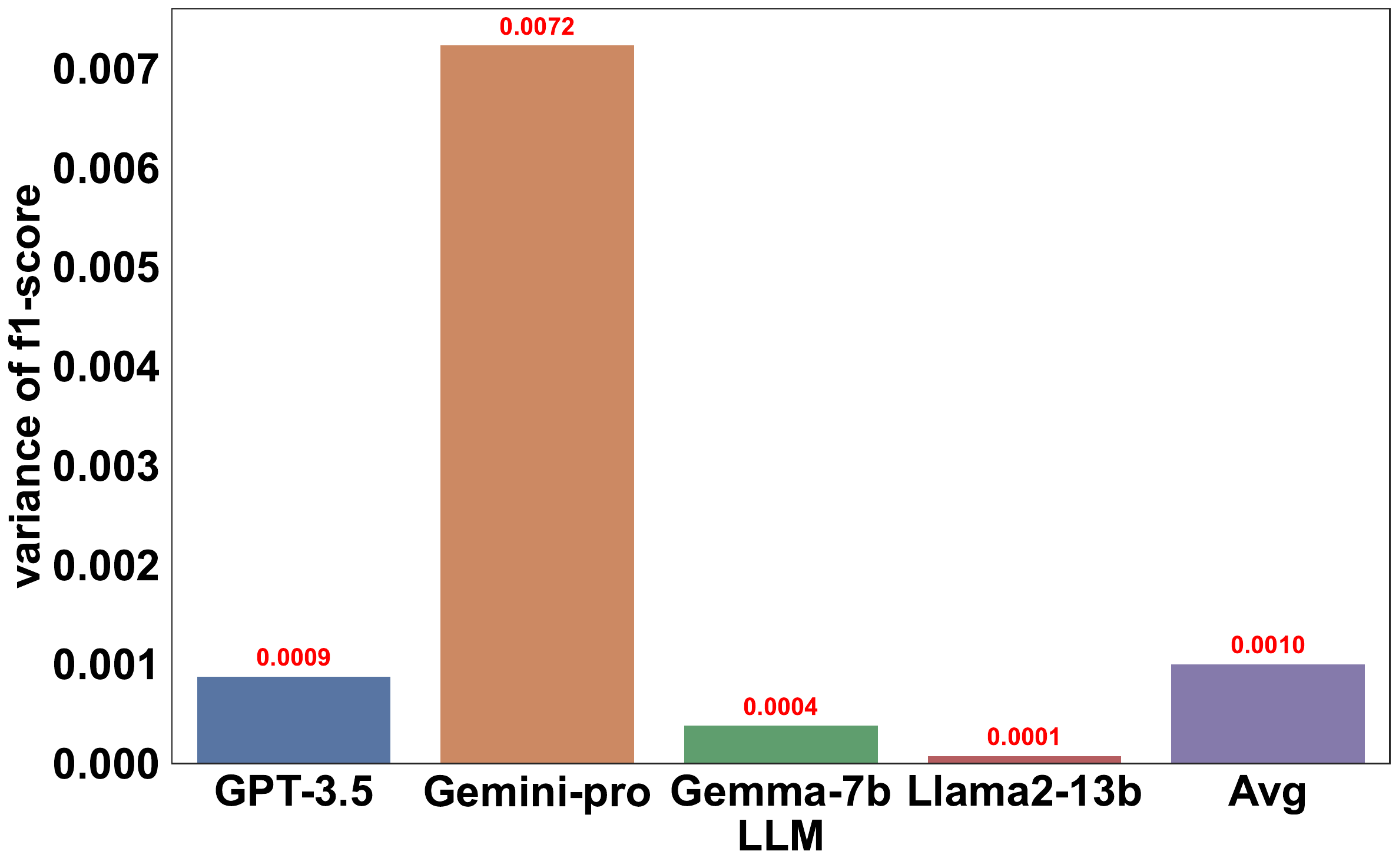}
    \caption{$F_1$-score for different languages on benign questions.}
    \label{fig:f1var}
\end{figure}
\begin{figure*}[t]
    \includegraphics[width=0.9\textwidth]{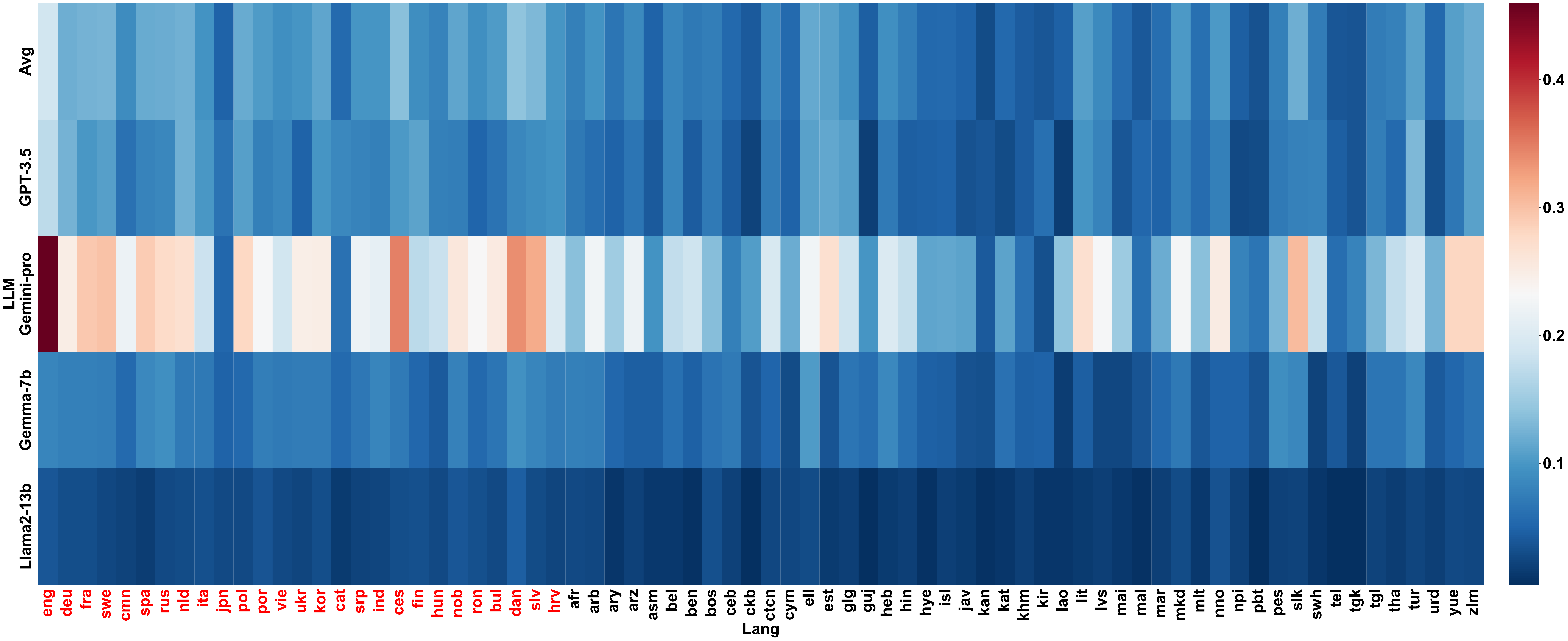}
    \caption{$F_1$-score for different languages on benign questions.}
    \label{fig:f1heat}
\end{figure*}
To answer this question, we exercise different LLMs with benign questions across various languages, and analyze the quality of their responses. Specifically, we first translate questions in NQ dataset to 74 languages, resulting in 2220 queries. Subsequently, we input these queries to four LLMs separately. For analytical purposes, we translate all non-English responses into English and assess the quality of each response.

Following related works~\cite{helm,nq}, we adopt $F_1$-score to comprehensively access the response quality. $F_1$-score is based on the precision and recall, which are calculated as follows:
\begin{align}
    F_1 &= \frac{2 \times precision \times recall}{precision + recall} \label{eq:f1} \\
    precision &= \frac{|\{x|x \in P ~ and ~ x \in \Lambda\}|}{|P|} \label{eq:pre}\\
    recall    &=\frac{|\{x|x \in P ~ and ~ x \in \Lambda\}|}{|\Lambda|} \label{eq:recall}
\end{align}
where $P$ denotes the word list of a response, $\Lambda$ denotes the word list of a reference answer, and $|\cdot|$ is the number of elements in the related set. Each question in the NQ dataset is accompanied by a list of short answers. It is necessary to perform specialized preprocessing on both the original response and the reference answers to ensure consistency and improve accuracy.
\begin{algorithm}[t]
    \caption{$f1(R, \Phi)$}
    \label{alg:f1}
    $\hat{R} \leftarrow$ perform preprocess to R\;
    $\hat{\Phi} \leftarrow$ perform preprocess to $\Phi$\;
    let $\Gamma$ be an empty set\;
    \For{each answer $\Lambda \in \hat{\Phi}$}{
        $s \leftarrow$ calculate $F_1$-score with $\hat{R}$ and $\Lambda$ according to Formula~\ref{eq:f1}-\ref{eq:recall}\;
        $\Gamma \leftarrow \Gamma \cup \{s\} $\;
    }
    \Return{$max(\Gamma)$}\;
\end{algorithm}

Algorithm~\ref{alg:f1} outlines the method for calculating the $F_1$-score given a response $R$ and a set of reference short answers $\Phi$. Initially, both $R$ and $\Phi$ are preprocessed by converting all words to lowercase, removing stop words, and eliminating special symbols such as periods (lines 1-2). This results in $\hat{R}$, a list of words, and $\hat{\Phi}$, a list of word lists corresponding to each short answer. Next, the algorithm computes the $F_1$-score for the response $\hat{R}$ against each short answer $\Gamma$ (lines 4-6). Ultimately, the maximum score across all comparisons with $\Gamma$ is returned as the final $F_1$-score of response $R$.

Figure~\ref{fig:f1var} shows the variance of $F_1$-score of each language for each LLM. The overall variance of $F_1$-score over 74 languages is 0.001. Surprisingly, Gemini-pro, despite being the most consistent in terms of safety adherence (with the lowest variance of LJR), exhibits the highest variance of $F_1$-score, indicating that the quality of responses provided by Gemini-pro varies significantly across different languages. We highlight that the high variance of $F_1$-score does not necessarily indicate a tendency for a LLM to produce low-quality responses. The heatmap of $F_1$-score in Figure~\ref{fig:f1heat} reveals that Gemini-pro exhibits a broad range of $F_1$-score values. Notably, Gemini-pro achieves higher $F_1$-scores in most languages compared to other LLMs, suggesting that it generally provides high-quality responses across various languages. In addition, both GPT-3.5 and Gemini-pro, as expected, have the best $F_1$-score on English (0.1723 and 0.46 respectively). However, Gemma-7b and Llama2-13b attain their highest $F_1$-scores in languages Greek (ell, 0.1038) and Danish (dan, 0.0449) respectively, instead of English. We will explain the possible reasons in RQ3. 
\begin{framed}
    \noindent \emph{Answer to RQ2: 
    All four LLMs tend to provide high-quality responses in high-resource languages but struggle to maintain the same level of quality across all languages.}
\end{framed}
\noindent\emph{\textbf{RQ3: Which languages do LLMs perform better?}} To answer this question, we introduce a new metric named comprehensive index (CI) by combining LJR and $F_1$-score. CI measures the overall performance of an LLM on a specific language. We defined CI as follows.
\begin{align}
    CI = \alpha \cdot F_1 - \beta \cdot LJR
    \label{fm:ci}
\end{align}
where $\alpha$ and $\beta$ can take values between 0 and 1. They control the significance of $F_1$-score and LJR, respectively. For example, if we prioritize the importance of safety performance, we can set a larger $\beta$. In this work, both $\alpha$ and $\beta$ are set to 0.5. Note that LJT and $F_1$ in Formula~\ref{fm:ci} are normalized values.
\begin{figure*}[t]
    \includegraphics[width=0.9\textwidth]{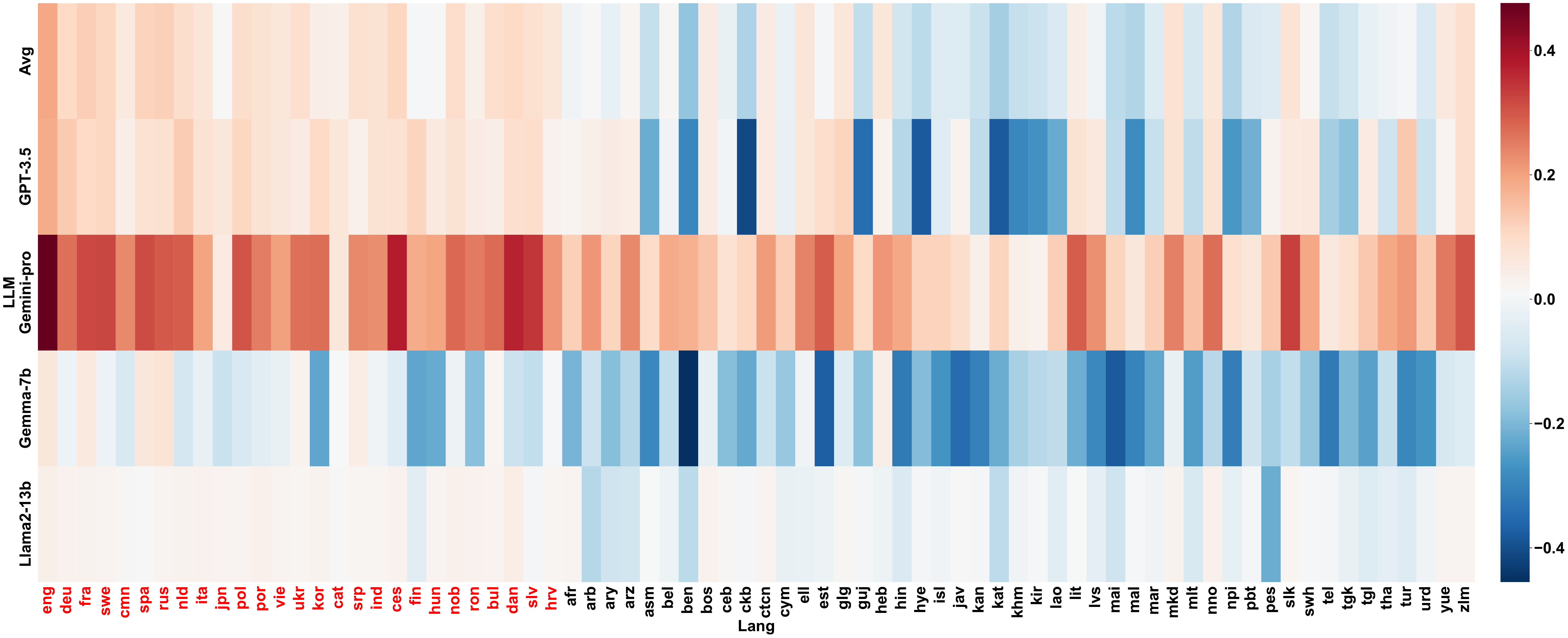}
    \caption{CI-score for different languages on four LLMs.}
    \label{fig:ciheat}
\end{figure*}
As expected, high-resource languages (highlighted in red) have higher CI scores. The top five languages with the highest CI scores are English (eng, 0.1899), French (fra, 0.1262), Russian (rus, 0.1205), Spanish (spa, 0.1181), and Czech (ces, 0.1103). In contrast, the bottom five languages are Bengali (ben, -0.1718), Georgian (kat, -0.1491), Central Kurdish (ckb, -0.1319), Nepali (npi, -0.1296), and Malayalam (mal, -0.1286). The significant discrepancies in comprehensive performance across different languages are noteworthy. Among all languages, English performs exceptionally well compared to other languages, with a CI score approximately 50\% higher than that of French, i.e, the second-highest language.

For individual LLMs, both GPT-3.5 and Gemini-pro perform best in English, with CI scores of 0.1847 and 0.4762, respectively. In a surprising outcome, the highest CI score for Llama2-13b is in Danish (dan,0.0439), while for Gemma-7b, it is in Russian (rus, 0.0733). English ranks as the second-highest language for Llama2-13b with a CI score of 0.037, and third-highest for Gemma-7b with a CI score of 0.0619.

Upon further analysis, we find that for both LLMs, the $F_1$-score in English is slightly lower than that in the highest language (0.07/0.09 for Llama2-13b and 0.17/0.19 for Gemma-7b). However, the LJR values are consistent across both languages (0.0/0.0 for Llama2-13b and 0.05/0.05 for Gemma-7b). The lower $F_1$-score in English may be due to lower precision, as English responses tend to provide more detailed answers. This assumption is supported by findings from Gemma-7b's responses. Among questions with the same recall in both English and Russian, there were five questions where Russian responses have higher precision compared to English, and only one question where the reverse is true. For example, in response to the question ``What are the names of the atom bombs dropped on Japan'', the English answer provides more detail: ``The two atom bombs dropped on Japan on August 6 and 9, 1945, were named 'Little Boy' and 'Fat Man' '', while the Russian answer is more concise: ``The atomic bombs dropped on Japan were called 'Little Boy' and 'Fat Man' ''.

\begin{framed}
    \noindent \emph{Answer to RQ3: 
    On average, English has the best comprehensive performance, which is 210\% higher than that of Bengali (which has the lowest CI) in terms of CI.}
\end{framed}

\section{Mitigation of linguistic discrimination}
\label{sec:4}
Based on the above findings, in this section, we introduce LDFighter, a lightweight approach designed to mitigate linguistic discrimination from both safety and quality perspectives. LDFighter involves several steps. First, a query is translated into a selection of $K$ languages. Then, each translated query is inputted into a target LLM to generate responses in different languages. Next, these responses are translated to English, and the average similarity of each response to others are calculated. Finally, the response with the highest average similarity is selected as the final answer to the original query. Note that the selected answer will be translated back into the language of the original query if required. In the following, we introduce how the similarity-based voting works for the response, and then assess the effectiveness of LDFighter based on three research questions. 

\subsection{Similarity-based Voting}
Intuitively, when presented with the same question, different individuals tend to offer answers that exhibit certain similarities (assuming they possess similar ground-truth knowledge). Inspired by this observation, we select the response most similar to others from a set of responses as the final output of the target LLM. To achieve this, we adopt a cosine similarity to measure the similarity between pairs of responses. 

Algorithm~\ref{alg:sm} details the process for generating a response using similarity-based voting. Given a set of responses $\mathcal{R}$ in English translated from different languages, each response is first encoded into a vector, resulting in a set of vectors $\mathcal{V}$ (line 1). Next, the average similarity of each vector to the other vectors is calculated according to Formula~\ref{fm:sm} (lines 2-5), and the vector with the highest average similarity is chosen (line 6). Finally, the response corresponding to the selected vector is returned as the final response of the LLM.
\begin{algorithm}[t]
    \caption{$vote(\mathcal{R})$}
    \label{alg:sm}
    $\mathcal{V} \leftarrow$ encode each response $r\in R$\;
    let $\Gamma$ be an empty set\;
    \For{each answer $v_k \in \mathcal{V}$}{
        $s \leftarrow avgCos(\mathcal{V}, v_k)$ \;
        $\Gamma \leftarrow \Gamma \cup \{s\} $\;
    }
    $\hat{v} \leftarrow max(\Gamma)$\;
    $r \leftarrow$ response responding to $\hat{v}$\;
    \Return{r}\;
\end{algorithm}

\begin{align}
    avgCos(\mathcal{V},v_k) = \frac{\sum_{j\neq k}Cos(v_k,v_j)}{|\mathcal{V}|-1}
    \label{fm:sm}
\end{align}
\subsection{Research Questions}
In the following, we evaluate the effectiveness of LDFighter by addressing three key research questions.
\begin{itemize}
    \item RQ4: How effective is LDFighter on improving the safety performance of LLMs across different languages?
    \item RQ5: How effective is LDFighter on improving the response quality of LLMs across different languages?
    \item RQ6: What is the cost of LDFighter?
\end{itemize}
LLMs equipped with LDFighter inherently offer consistent service across all languages. In addition to ensuring consistency, we investigate whether LDFighter can maintain or surpass the original performance of LLMs in terms of safety and quality. Therefore, RQ4 and RQ5 aim to assess LDFighter's effectiveness in improving multilingual safety performance and response quality, respectively. RQ6 evaluates the time efficiency of LDFighter. We adopt the state-of-the-art text embedding model SFR-Embedding-Mistral~\cite{SFRAIResearch2024,muennighoff2022mteb} to encode each response. All the experiments are conducted on a cloud server equipped with two RTX 4090D (24GB), a 36 vCPU AMD EPYC 9754 128-core processor and 120 GB RAM, running on Ubuntu 20.04.5 LTS(64 bit).
\subsection{Results}
\begin{figure*}[t] 
    \centering
    \begin{minipage}{0.48\textwidth}
        \centering
        \includegraphics[width=\textwidth]{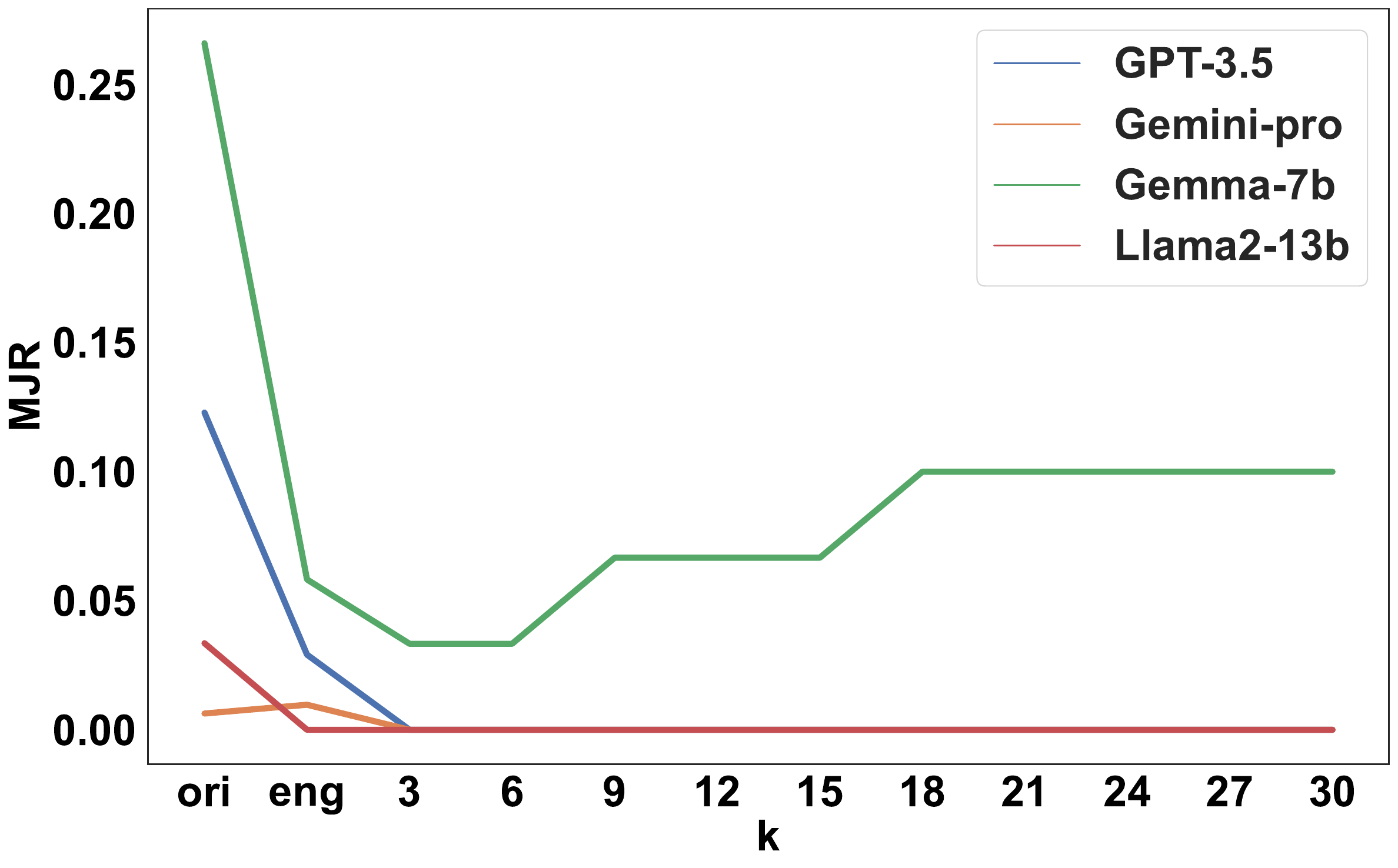}
        \caption{MJR of different LLMs with LDFighter}
        \label{fig:mjr_vote}
    \end{minipage} 
    \hfill
    \begin{minipage}{0.48\textwidth}
        \centering
        \includegraphics[width=\textwidth]{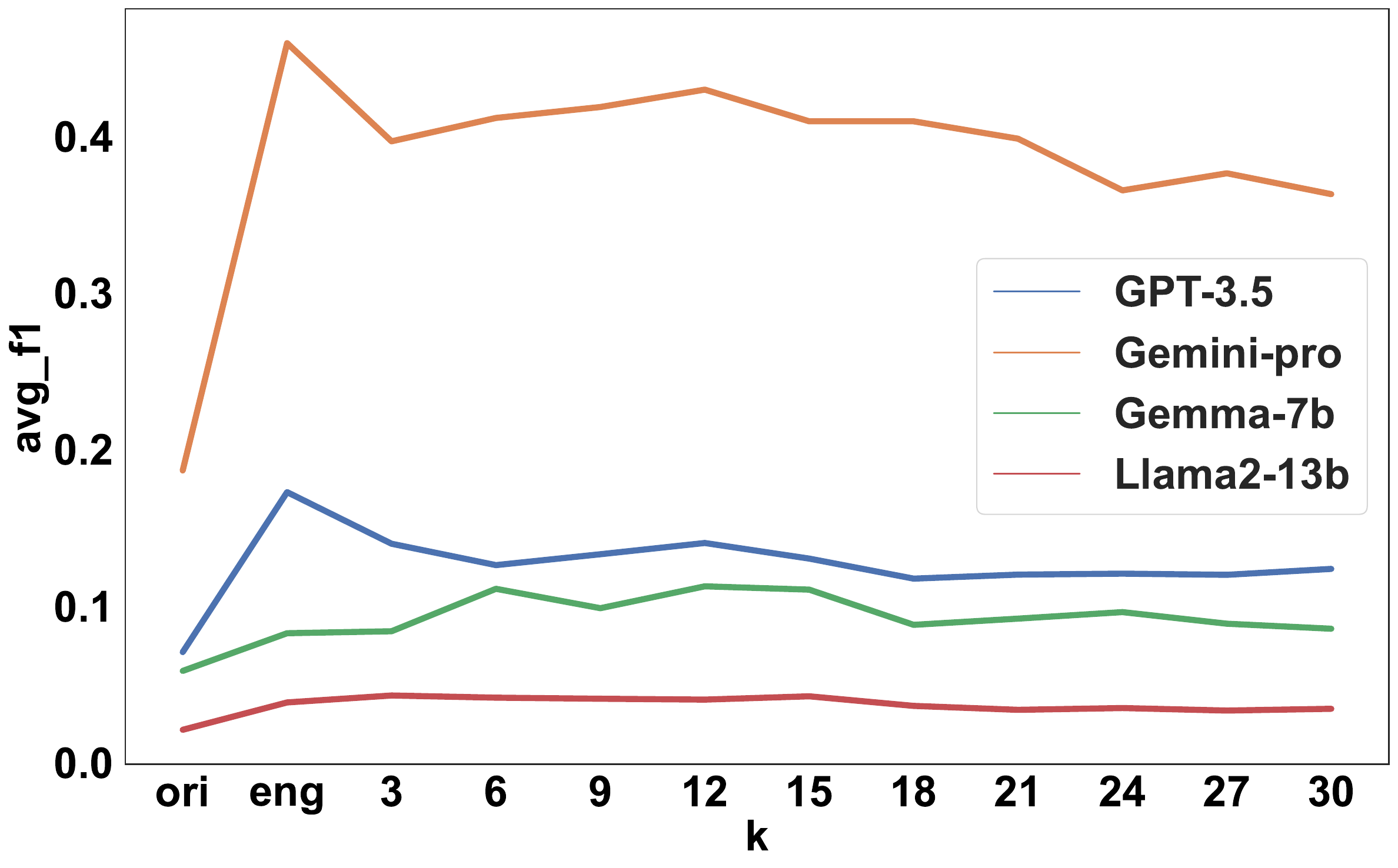} 
        \caption{$F_1$-score of different LLMs with LDFighter}
        \label{fig:f1_vote}
    \end{minipage}
\end{figure*}
\noindent\emph{\textbf{RQ4: How effective is LDFighter on improving the safety performance of LLMs across different languages?}} To answer this question, we apply LDFighter to AdvBench and report the MJR of each LLMs. For each LLM, we select the top three languages with the highest CI scores. To investigate the impact of the number of languages on MJR, we vary the number of languages $K$ from 3 to 30 with step 3. Additionally, we examine the performance of LDFighter when using only one language, i.e., English, to determine if using multiple languages is necessary. 

Figure~\ref{fig:mjr_vote} shows the MJR of four LLMs without or with the help of LDFighter. The values responding to ``ori'' on x-axis are the result of each LLM without LDFighter, the values responding to ``eng'' are the result of each LLM with LDFighter where only one language, i.e., English, is used. When the value of $k$ is set to be 3 or above, the MJR of each LLM drops significantly compared to the original MJR.  Particularly, the MJR of GPT-3.5, Gemini-pro and Llama2-13b falls straight to 0.0 when using the top three languages, and then remain unchanged as $k$ increases. For Gemma-7b, the MJR first decreases to the lowest point at $k=3$, then rises slightly and stabilizes around 0.1 after $k=18$. When using only English in LDFighter, only Llama2-13b achieves an MJR of 0.0, while the other three LLMs do not reach their optimal MJR. 

\begin{table}[t]
\caption{ASR of GCG adversarial samples on LDFighter-equipped Llama2-7b}
\label{tab:asr}
\begin{tabular}{lllllllll}
\toprule
Top-k & 3 & 4 & 5 & 6 & 7 & 8 & 9 & 10 \\
ASR   & 13\% & 8\% & 8\% & 6\% & 5\% & 4\% & 2\% & 2\% \\
\bottomrule
\end{tabular}
\end{table}
In addition to addressing inherent multilingual challenges, we further conduct a simple experiment to assess LDFighter's effectiveness in defending against state-of-the-art jailbreaking attacks. Specifically, we randomly selected adversarial prompts targeting Llama2-7b from the dataset created by CASPER~\cite{CASPER}, generated using the GCG approach~\cite{advbench}. Subsequently, LDFighter is applied to Llama2-7b to counter these prompts. To determine the optimal language selection strategy, we consider the top-k languages based on their average CI score as revealed in RQ3. Varying k from 3 to 10, we record the corresponding attack success rates (ASR), as detailed in Table~\ref{tab:asr}. Note that all selected adversarial prompts are in English and validated, resulting in an initial ASR of 100\% on Llama2-7b. Notably, the integration of LDFighter leads to a significant reduction in ASR. Even with just the top 3 languages (including English), the ASR plummets to 13\% from the baseline 100\%. Furthermore, utilizing the top 9 languages, the ASR drops to a mere 2\%. These results underscore the remarkable potential of LDFighter in thwarting state-of-the-art jailbreaking attacks.
Based on these experimental results, we can answer RQ4 as follows.
\begin{framed}
    \noindent \emph{Answer to RQ4: LDFighter demonstrates significant effectiveness in reducing the jailbreak rate of both harmful questions and adversarial prompts generated by GCG. Moreover, the use of three languages with the highest CI leads to the lowest MJR for three out of the four LLMs.} 
\end{framed}
\noindent\emph{\textbf{RQ5: How effective is LDFighter on improving the response quality of LLMs across different languages?}} To answer this question, we apply LDFighter to the NQ dataset and evaluate the average $F_1$-score of four LLMs. The top-k settings are consistent with RQ4.

Figure~\ref{fig:f1_vote} illustrates the average $F_1$-score of four LLMs, both with and without LDFighter. Compared to the original average $F_1$-score, all LLMs experience an increase in average $F_1$-score with LDFighter, though the extent varies. For GPT-3.5 and Gemini-pro, the average $F_1$-score rises significantly and peaks when using only English in LDFighter. As $k$ continues to increase, the average $F_1$-score slightly decreases but still remains higher than the original average. For Gemma-7b and Llama2-13b, the average $F_1$-score steadily improves at first, reaching its maximum at $k=12$ for Gemma-7b and K=3 for Llama2-13b, after which it stabilizes. We thus have the following answer to RQ5.
\begin{framed}
    \noindent \emph{Answer to RQ5: LDFighter substantially enhances the average $F_1$-score of responses across four LLMs, indicating its potential to improve response quality for speakers of different languages.}
\end{framed}
\noindent\emph{\textbf{RQ6: What is the time overhead of LDFighter?}} The time overhead of LDFighter primarily consists of three parts: translation, querying, and encoding. Compared to the time spent interacting with LLMs, the time required for similarity-based voting can be considered negligible if the embeddings of each response have already been prepared. Let $t_{tra}$, $t_{qry}$, and $t_{emd}$ represent the time costs of translation, querying with an LLM, and sentence encoding, respectively. Therefore, the overall time cost of applying LDFighter can be estimated as follows:
\begin{align}
    cost = 2k \cdot t_{tra}+k\cdot t_{qry} + k \cdot t_{emd}
\end{align}
where $k$ is the number of languages used in LDFighter. The translation time is $2k \cdot t_{tra}$ because we need to translate both the original query and responses in different languages.

The time overhead varies linearly with the value of $k$, but it can be reduced by parallelizing the computation process of $k$ languages. We thus report the cost of LDFighter with k=1. On average, the time cost per query is 9.96 seconds on GPT-3.5, 4.91 seconds on Gemini-pro, 9.68 seconds on Gemma-7b, and 11.75 seconds on Llama2-13b. We thus have the following answer to RQ6.
\begin{framed}
    \noindent \emph{Answer to RQ6: LDFighter has the potential to be applied at runtime.}
\end{framed}
We remark that indeed there is additional token-cost when close-source API such as ChatGPT is adopted. However, such cost can be effectively mitigated if we adopt LDFighter with only English. 
\section{Related work}
\label{sec:5}
This work is related to multilingual LLM jailbreak. Puttaparthi et al.~\cite{puttaparthi2023comprehensive} explore the reliability of ChatGPT for multilingual queries, assessing its jailbreak rate across 30 malicious questions and 121 languages in four scenarios: single language, mixed languages, responses in a different language, and multilingual wrapping. Yong et al.~\cite{yong2023low} analyze GPT-4's safety performance across 12 low-resource languages, and find that GPT-4 fails to generalize its safety mechanisms to low-resource languages. Li et al.~\cite{liJailbreak2024} conduct an empirical study of multilingual LLM jailbreak attacks. They create a multilingual jailbreak dataset to assess the safety performance of four LLMs across nine languages. They also examine attention distribution on failed and successful jailbreak cases and reveal that LLMs have a more balanced attention distribution on failed jailbreak cases. Deng et al.~\cite{deng2024multilingual} investigate the multilingual jailbreak of GPT-3.5 and GPT-4 across 30 languages under both unintentional and intentional scenarios. In the unintentional scenario, input questions are not deliberately modified or wrapped with malicious prompt, while in the intentional scenario, input questions are carefully designed to bypass the safety mechanisms. Previous research has either been limited in language coverage or focused primarily on the GPT series. In contrast, our study examines a broader range of languages and incorporates both state-of-the-art closed and open-source LLMs.

This work is related to the defense of LLM jailbreak. Robey et al.~\cite{robey2023smoothllm} propose a defense against the 
GCG attack~\cite{advbench}. This approach begins by creating a set of mutated inputs through character-level perturbations of the original input. Such perturbation may render the adversarial suffixes invalid. After that, the approach decides the input whether is harmful by majority voting on responses of the mutated inputs. If more than half of responses are refusals, the input is then classified as harmful. However, this approach may not be effective against natural jailbreak questions generated by translation-based attacks. To tack the multilingual jailbreak threats, Deng et al.\cite{deng2024multilingual} and Li et al.~\cite{liJailbreak2024} propose to fine-tune LLMs with a set of multilingual input-output pairs that encompass both unsafe and general query questions. Although effective, fine-tuning LLMs can be time-consuming and complex. In contrast, LDFighter offers a convenient alternative that does not require modifying the LLM itself and can be easily implemented as a plug-and-play solution.

This work is also related to the multilingual discrimination in natural language processing. Blasi et al.~\cite{blasi2021systematic} systematically investigate multilingual inequalities in language technologies, examining user-facing technologies such as question answering, as well as foundational NLP tasks like dependency parsing across the world's languages. Their study reveals significant disparities in the development of language technologies across different languages. Sourojit Ghosh and Aylin Caliskan~\cite{ghosh2023chatgpt} evaluating ChatGPT's translation performance in several low-resource languages and discovers that ChatGPT tends to perpetuate gender defaults and stereotypes associated with certain occupations. We remark that our work focuses on a wide spectrum of LLMs and evaluates their multilingual discrimination from both safety and quality aspects. 
\section{Conclusion}
\label{sec:6}
In this work, we evaluate the consistency of LLM outputs in response to queries in various languages from two aspects, focusing on safety and quality. Through experiments involving four state-of-the-art LLMs and 74 languages, we identify significant linguistic discrepancies in LLM performance across different languages. To address these issues, we introduce LDFighter, a lightweight approach that enhances the safety and quality of LLM responses while ensuring consistent service for speakers of all languages.

\bibliographystyle{ACM-Reference-Format}
\bibliography{ref}


\begin{thebibliography}{36}


\ifx \showCODEN    \undefined \def \showCODEN     #1{\unskip}     \fi
\ifx \showDOI      \undefined \def \showDOI       #1{#1}\fi
\ifx \showISBNx    \undefined \def \showISBNx     #1{\unskip}     \fi
\ifx \showISBNxiii \undefined \def \showISBNxiii  #1{\unskip}     \fi
\ifx \showISSN     \undefined \def \showISSN      #1{\unskip}     \fi
\ifx \showLCCN     \undefined \def \showLCCN      #1{\unskip}     \fi
\ifx \shownote     \undefined \def \shownote      #1{#1}          \fi
\ifx \showarticletitle \undefined \def \showarticletitle #1{#1}   \fi
\ifx \showURL      \undefined \def \showURL       {\relax}        \fi
\providecommand\bibfield[2]{#2}
\providecommand\bibinfo[2]{#2}
\providecommand\natexlab[1]{#1}
\providecommand\showeprint[2][]{arXiv:#2}

\bibitem[Barrault et~al\mbox{.}(2023)]%
        {seamless2023}
\bibfield{author}{\bibinfo{person}{Lo{\"\i}c Barrault}, \bibinfo{person}{Yu-An
  Chung}, \bibinfo{person}{Mariano~Coria Meglioli}, \bibinfo{person}{David
  Dale}, \bibinfo{person}{Ning Dong}, \bibinfo{person}{Mark Duppenthaler},
  \bibinfo{person}{Paul-Ambroise Duquenne}, \bibinfo{person}{Brian Ellis},
  \bibinfo{person}{Hady Elsahar}, \bibinfo{person}{Justin Haaheim},
  {et~al\mbox{.}}} \bibinfo{year}{2023}\natexlab{}.
\newblock \showarticletitle{Seamless: Multilingual Expressive and Streaming
  Speech Translation}.
\newblock \bibinfo{journal}{\emph{arXiv preprint arXiv:2312.05187}}
  (\bibinfo{year}{2023}).
\newblock


\bibitem[Blasi et~al\mbox{.}({[n.\,d.]})]%
        {blasi2022}
\bibfield{author}{\bibinfo{person}{Damian Blasi}, \bibinfo{person}{Antonios
  Anastasopoulos}, {and} \bibinfo{person}{Graham Neubig}.}
  \bibinfo{year}{[n.\,d.]}\natexlab{}.
\newblock \showarticletitle{Systematic Inequalities in Language Technology
  Performance across the World's Languages}. In
  \bibinfo{booktitle}{\emph{Proceedings of the 60th Annual Meeting of the
  Association for Computational Linguistics (Volume 1: Long Papers)}} (Dublin,
  Ireland, 2022-05), \bibfield{editor}{\bibinfo{person}{Smaranda Muresan},
  \bibinfo{person}{Preslav Nakov}, {and} \bibinfo{person}{Aline Villavicencio}}
  (Eds.). \bibinfo{publisher}{Association for Computational Linguistics},
  \bibinfo{pages}{5486--5505}.
\newblock
\urldef\tempurl%
\url{https://doi.org/10.18653/v1/2022.acl-long.376}
\showDOI{\tempurl}


\bibitem[Blasi et~al\mbox{.}(2021)]%
        {blasi2021systematic}
\bibfield{author}{\bibinfo{person}{Dami{\'a}n Blasi}, \bibinfo{person}{Antonios
  Anastasopoulos}, {and} \bibinfo{person}{Graham Neubig}.}
  \bibinfo{year}{2021}\natexlab{}.
\newblock \showarticletitle{Systematic Inequalities in Language Technology
  Performance across the World's Languages}.
\newblock \bibinfo{journal}{\emph{arXiv preprint arXiv:2110.06733}}
  (\bibinfo{year}{2021}).
\newblock


\bibitem[Brown et~al\mbox{.}(2020)]%
        {gpt3}
\bibfield{author}{\bibinfo{person}{Tom Brown}, \bibinfo{person}{Benjamin Mann},
  \bibinfo{person}{Nick Ryder}, \bibinfo{person}{Melanie Subbiah},
  \bibinfo{person}{Jared~D Kaplan}, \bibinfo{person}{Prafulla Dhariwal},
  \bibinfo{person}{Arvind Neelakantan}, \bibinfo{person}{Pranav Shyam},
  \bibinfo{person}{Girish Sastry}, \bibinfo{person}{Amanda Askell},
  {et~al\mbox{.}}} \bibinfo{year}{2020}\natexlab{}.
\newblock \showarticletitle{Language models are few-shot learners}.
\newblock \bibinfo{journal}{\emph{Advances in neural information processing
  systems}}  \bibinfo{volume}{33} (\bibinfo{year}{2020}),
  \bibinfo{pages}{1877--1901}.
\newblock


\bibitem[Chang et~al\mbox{.}({[n.\,d.]})]%
        {chang2023}
\bibfield{author}{\bibinfo{person}{Tyler~A. Chang}, \bibinfo{person}{Catherine
  Arnett}, \bibinfo{person}{Zhuowen Tu}, {and} \bibinfo{person}{Benjamin~K.
  Bergen}.} \bibinfo{year}{[n.\,d.]}\natexlab{}.
\newblock \bibinfo{booktitle}{\emph{When Is Multilinguality a Curse? Language
  Modeling for 250 High- and Low-Resource Languages}}.
\newblock
\showeprint[arxiv]{2311.09205}~[cs]
\urldef\tempurl%
\url{http://arxiv.org/abs/2311.09205}
\showURL{%
\tempurl}


\bibitem[Chang et~al\mbox{.}(2023)]%
        {chang2023survey}
\bibfield{author}{\bibinfo{person}{Yupeng Chang}, \bibinfo{person}{Xu Wang},
  \bibinfo{person}{Jindong Wang}, \bibinfo{person}{Yuan Wu},
  \bibinfo{person}{Linyi Yang}, \bibinfo{person}{Kaijie Zhu},
  \bibinfo{person}{Hao Chen}, \bibinfo{person}{Xiaoyuan Yi},
  \bibinfo{person}{Cunxiang Wang}, \bibinfo{person}{Yidong Wang},
  {et~al\mbox{.}}} \bibinfo{year}{2023}\natexlab{}.
\newblock \showarticletitle{A survey on evaluation of large language models}.
\newblock \bibinfo{journal}{\emph{ACM Transactions on Intelligent Systems and
  Technology}} (\bibinfo{year}{2023}).
\newblock


\bibitem[Deng et~al\mbox{.}(2024)]%
        {deng2024multilingual}
\bibfield{author}{\bibinfo{person}{Yue Deng}, \bibinfo{person}{Wenxuan Zhang},
  \bibinfo{person}{Sinno~Jialin Pan}, {and} \bibinfo{person}{Lidong Bing}.}
  \bibinfo{year}{2024}\natexlab{}.
\newblock \showarticletitle{Multilingual Jailbreak Challenges in Large Language
  Models}. In \bibinfo{booktitle}{\emph{The Twelfth International Conference on
  Learning Representations}}.
\newblock
\urldef\tempurl%
\url{https://openreview.net/forum?id=vESNKdEMGp}
\showURL{%
\tempurl}


\bibitem[Dong et~al\mbox{.}(2022)]%
        {icl}
\bibfield{author}{\bibinfo{person}{Qingxiu Dong}, \bibinfo{person}{Lei Li},
  \bibinfo{person}{Damai Dai}, \bibinfo{person}{Ce Zheng},
  \bibinfo{person}{Zhiyong Wu}, \bibinfo{person}{Baobao Chang},
  \bibinfo{person}{Xu Sun}, \bibinfo{person}{Jingjing Xu}, {and}
  \bibinfo{person}{Zhifang Sui}.} \bibinfo{year}{2022}\natexlab{}.
\newblock \showarticletitle{A survey for in-context learning}.
\newblock \bibinfo{journal}{\emph{arXiv preprint arXiv:2301.00234}}
  (\bibinfo{year}{2022}).
\newblock


\bibitem[Ghosh and Caliskan(2023)]%
        {ghosh2023chatgpt}
\bibfield{author}{\bibinfo{person}{Sourojit Ghosh} {and} \bibinfo{person}{Aylin
  Caliskan}.} \bibinfo{year}{2023}\natexlab{}.
\newblock \showarticletitle{Chatgpt perpetuates gender bias in machine
  translation and ignores non-gendered pronouns: Findings across bengali and
  five other low-resource languages}. In \bibinfo{booktitle}{\emph{Proceedings
  of the 2023 AAAI/ACM Conference on AI, Ethics, and Society}}.
  \bibinfo{pages}{901--912}.
\newblock


\bibitem[Google({[n.\,d.]})]%
        {geminipro}
\bibfield{author}{\bibinfo{person}{Google}.}
  \bibinfo{year}{[n.\,d.]}\natexlab{}.
\newblock
  \bibinfo{howpublished}{\url{https://ai.google.dev/?gad_source=1&gclid=EAIaIQobChMI7tit_Mm6hQMVsqhmAh0gygCvEAAYASAAEgKD4_D_BwE}}.
\newblock


\bibitem[Hendy et~al\mbox{.}(2023)]%
        {hendy2023good}
\bibfield{author}{\bibinfo{person}{Amr Hendy}, \bibinfo{person}{Mohamed
  Abdelrehim}, \bibinfo{person}{Amr Sharaf}, \bibinfo{person}{Vikas Raunak},
  \bibinfo{person}{Mohamed Gabr}, \bibinfo{person}{Hitokazu Matsushita},
  \bibinfo{person}{Young~Jin Kim}, \bibinfo{person}{Mohamed Afify}, {and}
  \bibinfo{person}{Hany~Hassan Awadalla}.} \bibinfo{year}{2023}\natexlab{}.
\newblock \showarticletitle{How good are gpt models at machine translation? a
  comprehensive evaluation}.
\newblock \bibinfo{journal}{\emph{arXiv preprint arXiv:2302.09210}}
  (\bibinfo{year}{2023}).
\newblock


\bibitem[Kojima et~al\mbox{.}(2022)]%
        {kojima2022large}
\bibfield{author}{\bibinfo{person}{Takeshi Kojima},
  \bibinfo{person}{Shixiang~Shane Gu}, \bibinfo{person}{Machel Reid},
  \bibinfo{person}{Yutaka Matsuo}, {and} \bibinfo{person}{Yusuke Iwasawa}.}
  \bibinfo{year}{2022}\natexlab{}.
\newblock \showarticletitle{Large language models are zero-shot reasoners}.
\newblock \bibinfo{journal}{\emph{Advances in neural information processing
  systems}}  \bibinfo{volume}{35} (\bibinfo{year}{2022}),
  \bibinfo{pages}{22199--22213}.
\newblock


\bibitem[Kwiatkowski et~al\mbox{.}(2019)]%
        {nq}
\bibfield{author}{\bibinfo{person}{Tom Kwiatkowski},
  \bibinfo{person}{Jennimaria Palomaki}, \bibinfo{person}{Olivia Redfield},
  \bibinfo{person}{Michael Collins}, \bibinfo{person}{Ankur Parikh},
  \bibinfo{person}{Chris Alberti}, \bibinfo{person}{Danielle Epstein},
  \bibinfo{person}{Illia Polosukhin}, \bibinfo{person}{Jacob Devlin},
  \bibinfo{person}{Kenton Lee}, \bibinfo{person}{Kristina Toutanova},
  \bibinfo{person}{Llion Jones}, \bibinfo{person}{Matthew Kelcey},
  \bibinfo{person}{Ming-Wei Chang}, \bibinfo{person}{Andrew~M. Dai},
  \bibinfo{person}{Jakob Uszkoreit}, \bibinfo{person}{Quoc Le}, {and}
  \bibinfo{person}{Slav Petrov}.} \bibinfo{year}{2019}\natexlab{}.
\newblock \showarticletitle{Natural Questions: A Benchmark for Question
  Answering Research}.
\newblock \bibinfo{journal}{\emph{Transactions of the Association for
  Computational Linguistics}}  \bibinfo{volume}{7} (\bibinfo{year}{2019}),
  \bibinfo{pages}{452--466}.
\newblock
\urldef\tempurl%
\url{https://doi.org/10.1162/tacl_a_00276}
\showDOI{\tempurl}


\bibitem[Lankford et~al\mbox{.}({[n.\,d.]})]%
        {lankford2023}
\bibfield{author}{\bibinfo{person}{Séamus Lankford}, \bibinfo{person}{Haithem
  Afli}, {and} \bibinfo{person}{Andy Way}.}
  \bibinfo{year}{[n.\,d.]}\natexlab{}.
\newblock \showarticletitle{adaptMLLM: Fine-Tuning Multilingual Language Models
  on Low-Resource Languages with Integrated LLM Playgrounds}.
\newblock  \bibinfo{volume}{14}, \bibinfo{number}{12}
  (\bibinfo{year}{[n.\,d.]}), \bibinfo{pages}{638}.
\newblock
\showISSN{2078-2489}
\urldef\tempurl%
\url{https://doi.org/10.3390/info14120638}
\showDOI{\tempurl}


\bibitem[Li et~al\mbox{.}(2024)]%
        {liJailbreak2024}
\bibfield{author}{\bibinfo{person}{Jie Li}, \bibinfo{person}{Yi Liu},
  \bibinfo{person}{Chongyang Liu}, \bibinfo{person}{Ling Shi},
  \bibinfo{person}{Xiaoning Ren}, \bibinfo{person}{Yaowen Zheng},
  \bibinfo{person}{Yang Liu}, {and} \bibinfo{person}{Yinxing Xue}.}
  \bibinfo{year}{2024}\natexlab{}.
\newblock \bibinfo{title}{A {{Cross-Language Investigation}} into {{Jailbreak
  Attacks}} in {{Large Language Models}}}.
\newblock
\newblock
\showeprint{2401.16765}~[cs]


\bibitem[Liang et~al\mbox{.}(2022)]%
        {helm}
\bibfield{author}{\bibinfo{person}{Percy Liang}, \bibinfo{person}{Rishi
  Bommasani}, \bibinfo{person}{Tony Lee}, \bibinfo{person}{Dimitris Tsipras},
  \bibinfo{person}{Dilara Soylu}, \bibinfo{person}{Michihiro Yasunaga},
  \bibinfo{person}{Yian Zhang}, \bibinfo{person}{Deepak Narayanan},
  \bibinfo{person}{Yuhuai Wu}, \bibinfo{person}{Ananya Kumar}, {et~al\mbox{.}}}
  \bibinfo{year}{2022}\natexlab{}.
\newblock \showarticletitle{Holistic evaluation of language models}.
\newblock \bibinfo{journal}{\emph{arXiv preprint arXiv:2211.09110}}
  (\bibinfo{year}{2022}).
\newblock


\bibitem[Muennighoff et~al\mbox{.}(2022)]%
        {muennighoff2022mteb}
\bibfield{author}{\bibinfo{person}{Niklas Muennighoff},
  \bibinfo{person}{Nouamane Tazi}, \bibinfo{person}{Lo{\"\i}c Magne}, {and}
  \bibinfo{person}{Nils Reimers}.} \bibinfo{year}{2022}\natexlab{}.
\newblock \showarticletitle{MTEB: Massive Text Embedding Benchmark}.
\newblock \bibinfo{journal}{\emph{arXiv preprint arXiv:2210.07316}}
  (\bibinfo{year}{2022}).
\newblock
\urldef\tempurl%
\url{https://doi.org/10.48550/ARXIV.2210.07316}
\showDOI{\tempurl}


\bibitem[OpenAI({[n.\,d.]})]%
        {chatgpt}
\bibfield{author}{\bibinfo{person}{OpenAI}.}
  \bibinfo{year}{[n.\,d.]}\natexlab{}.
\newblock \bibinfo{howpublished}{\url{https://chat.openai.com/}}.
\newblock


\bibitem[OpenAI(2023)]%
        {gpt35}
\bibfield{author}{\bibinfo{person}{OpenAI}.} \bibinfo{year}{2023}\natexlab{}.
\newblock \bibinfo{howpublished}{\url{https://openai.com/product}}.
\newblock


\bibitem[Ouyang et~al\mbox{.}(2022)]%
        {ouyang2022training}
\bibfield{author}{\bibinfo{person}{Long Ouyang}, \bibinfo{person}{Jeffrey Wu},
  \bibinfo{person}{Xu Jiang}, \bibinfo{person}{Diogo Almeida},
  \bibinfo{person}{Carroll Wainwright}, \bibinfo{person}{Pamela Mishkin},
  \bibinfo{person}{Chong Zhang}, \bibinfo{person}{Sandhini Agarwal},
  \bibinfo{person}{Katarina Slama}, \bibinfo{person}{Alex Ray},
  {et~al\mbox{.}}} \bibinfo{year}{2022}\natexlab{}.
\newblock \showarticletitle{Training language models to follow instructions
  with human feedback}.
\newblock \bibinfo{journal}{\emph{Advances in neural information processing
  systems}}  \bibinfo{volume}{35} (\bibinfo{year}{2022}),
  \bibinfo{pages}{27730--27744}.
\newblock


\bibitem[Peng et~al\mbox{.}(2023)]%
        {peng2023towards}
\bibfield{author}{\bibinfo{person}{Keqin Peng}, \bibinfo{person}{Liang Ding},
  \bibinfo{person}{Qihuang Zhong}, \bibinfo{person}{Li Shen},
  \bibinfo{person}{Xuebo Liu}, \bibinfo{person}{Min Zhang},
  \bibinfo{person}{Yuanxin Ouyang}, {and} \bibinfo{person}{Dacheng Tao}.}
  \bibinfo{year}{2023}\natexlab{}.
\newblock \showarticletitle{Towards making the most of chatgpt for machine
  translation}.
\newblock \bibinfo{journal}{\emph{arXiv preprint arXiv:2303.13780}}
  (\bibinfo{year}{2023}).
\newblock


\bibitem[Puttaparthi et~al\mbox{.}(2023)]%
        {puttaparthi2023comprehensive}
\bibfield{author}{\bibinfo{person}{Poorna Chander~Reddy Puttaparthi},
  \bibinfo{person}{Soham~Sanjay Deo}, \bibinfo{person}{Hakan Gul},
  \bibinfo{person}{Yiming Tang}, \bibinfo{person}{Weiyi Shang}, {and}
  \bibinfo{person}{Zhe Yu}.} \bibinfo{year}{2023}\natexlab{}.
\newblock \showarticletitle{Comprehensive evaluation of chatgpt reliability
  through multilingual inquiries}.
\newblock \bibinfo{journal}{\emph{arXiv preprint arXiv:2312.10524}}
  (\bibinfo{year}{2023}).
\newblock


\bibitem[Radford et~al\mbox{.}(2019)]%
        {gpt2}
\bibfield{author}{\bibinfo{person}{Alec Radford}, \bibinfo{person}{Jeffrey Wu},
  \bibinfo{person}{Rewon Child}, \bibinfo{person}{David Luan},
  \bibinfo{person}{Dario Amodei}, \bibinfo{person}{Ilya Sutskever},
  {et~al\mbox{.}}} \bibinfo{year}{2019}\natexlab{}.
\newblock \showarticletitle{Language models are unsupervised multitask
  learners}.
\newblock \bibinfo{journal}{\emph{OpenAI blog}} \bibinfo{volume}{1},
  \bibinfo{number}{8} (\bibinfo{year}{2019}), \bibinfo{pages}{9}.
\newblock


\bibitem[Robey et~al\mbox{.}(2023)]%
        {robey2023smoothllm}
\bibfield{author}{\bibinfo{person}{Alexander Robey}, \bibinfo{person}{Eric
  Wong}, \bibinfo{person}{Hamed Hassani}, {and} \bibinfo{person}{George~J
  Pappas}.} \bibinfo{year}{2023}\natexlab{}.
\newblock \showarticletitle{Smoothllm: Defending large language models against
  jailbreaking attacks}.
\newblock \bibinfo{journal}{\emph{arXiv preprint arXiv:2310.03684}}
  (\bibinfo{year}{2023}).
\newblock


\bibitem[{Rui Meng and Ye Liu and Shafiq Rayhan Joty and Caiming Xiong and
  Yingbo Zhou and Semih Yavuz}(2024)]%
        {SFRAIResearch2024}
\bibfield{author}{\bibinfo{person}{{Rui Meng and Ye Liu and Shafiq Rayhan Joty
  and Caiming Xiong and Yingbo Zhou and Semih Yavuz}}.}
  \bibinfo{year}{2024}\natexlab{}.
\newblock \bibinfo{title}{SFR-Embedding-Mistral:Enhance Text Retrieval with
  Transfer Learning}.
\newblock \bibinfo{howpublished}{Salesforce AI Research Blog}.
\newblock
\urldef\tempurl%
\url{https://blog.salesforceairesearch.com/sfr-embedded-mistral/}
\showURL{%
\tempurl}


\bibitem[Tan et~al\mbox{.}(2023)]%
        {tan2023evaluation}
\bibfield{author}{\bibinfo{person}{Yiming Tan}, \bibinfo{person}{Dehai Min},
  \bibinfo{person}{Yu Li}, \bibinfo{person}{Wenbo Li}, \bibinfo{person}{Nan
  Hu}, \bibinfo{person}{Yongrui Chen}, {and} \bibinfo{person}{Guilin Qi}.}
  \bibinfo{year}{2023}\natexlab{}.
\newblock \showarticletitle{Evaluation of ChatGPT as a question answering
  system for answering complex questions}.
\newblock \bibinfo{journal}{\emph{arXiv preprint arXiv:2303.07992}}
  (\bibinfo{year}{2023}).
\newblock


\bibitem[Team et~al\mbox{.}(2024)]%
        {gemma}
\bibfield{author}{\bibinfo{person}{Gemma Team}, \bibinfo{person}{Thomas
  Mesnard}, \bibinfo{person}{Cassidy Hardin}, \bibinfo{person}{Robert Dadashi},
  \bibinfo{person}{Surya Bhupatiraju}, \bibinfo{person}{Shreya Pathak},
  \bibinfo{person}{Laurent Sifre}, \bibinfo{person}{Morgane Rivi{\`e}re},
  \bibinfo{person}{Mihir~Sanjay Kale}, \bibinfo{person}{Juliette Love},
  {et~al\mbox{.}}} \bibinfo{year}{2024}\natexlab{}.
\newblock \showarticletitle{Gemma: Open models based on gemini research and
  technology}.
\newblock \bibinfo{journal}{\emph{arXiv preprint arXiv:2403.08295}}
  (\bibinfo{year}{2024}).
\newblock


\bibitem[Team et~al\mbox{.}({[n.\,d.]})]%
        {teamNoLanguageLeft}
\bibfield{author}{\bibinfo{person}{{\relax NLLB} Team},
  \bibinfo{person}{Marta~R {Costa-juss{\`a}}}, \bibinfo{person}{James Cross},
  \bibinfo{person}{Onur {\c C}elebi}, \bibinfo{person}{Maha Elbayad},
  \bibinfo{person}{Kenneth Heafield}, \bibinfo{person}{Kevin Heffernan},
  \bibinfo{person}{Elahe Kalbassi}, \bibinfo{person}{Janice Lam},
  \bibinfo{person}{Daniel Licht}, \bibinfo{person}{Jean Maillard},
  \bibinfo{person}{Anna Sun}, \bibinfo{person}{Skyler Wang},
  \bibinfo{person}{Guillaume Wenzek}, \bibinfo{person}{Al Youngblood},
  \bibinfo{person}{Bapi Akula}, \bibinfo{person}{Loic Barrault},
  \bibinfo{person}{Gabriel~Mejia Gonzalez}, \bibinfo{person}{Prangthip
  Hansanti}, \bibinfo{person}{John Hoffman}, \bibinfo{person}{Semarley
  Jarrett}, \bibinfo{person}{Kaushik~Ram Sadagopan}, \bibinfo{person}{Dirk
  Rowe}, \bibinfo{person}{Shannon Spruit}, \bibinfo{person}{Chau Tran},
  \bibinfo{person}{Pierre Andrews}, \bibinfo{person}{Necip~Fazil Ayan},
  \bibinfo{person}{Shruti Bhosale}, \bibinfo{person}{Sergey Edunov},
  \bibinfo{person}{Angela Fan}, \bibinfo{person}{Cynthia Gao},
  \bibinfo{person}{Vedanuj Goswami}, \bibinfo{person}{Francisco Guzm{\'a}n},
  \bibinfo{person}{Philipp Koehn}, \bibinfo{person}{Alexandre Mourachko},
  \bibinfo{person}{Christophe Ropers}, \bibinfo{person}{Safiyyah Saleem},
  \bibinfo{person}{Holger Schwenk}, \bibinfo{person}{Jeff Wang}, {and}
  \bibinfo{person}{Meta Ai}.} \bibinfo{year}{[n.\,d.]}\natexlab{}.
\newblock \showarticletitle{No {{Language Left Behind}}: {{Scaling
  Human-Centered Machine Translation}}}.
\newblock  (\bibinfo{year}{[n.\,d.]}).
\newblock


\bibitem[Touvron et~al\mbox{.}({[n.\,d.]})]%
        {llama2}
\bibfield{author}{\bibinfo{person}{Hugo Touvron}, \bibinfo{person}{Louis
  Martin}, {and} \bibinfo{person}{Kevin Stone}.}
  \bibinfo{year}{[n.\,d.]}\natexlab{}.
\newblock \showarticletitle{Llama 2: Open Foundation and Fine-Tuned Chat
  Models}.
\newblock  (\bibinfo{year}{[n.\,d.]}).
\newblock


\bibitem[Wei et~al\mbox{.}(2022)]%
        {cot}
\bibfield{author}{\bibinfo{person}{Jason Wei}, \bibinfo{person}{Xuezhi Wang},
  \bibinfo{person}{Dale Schuurmans}, \bibinfo{person}{Maarten Bosma},
  \bibinfo{person}{Fei Xia}, \bibinfo{person}{Ed Chi}, \bibinfo{person}{Quoc~V
  Le}, \bibinfo{person}{Denny Zhou}, {et~al\mbox{.}}}
  \bibinfo{year}{2022}\natexlab{}.
\newblock \showarticletitle{Chain-of-thought prompting elicits reasoning in
  large language models}.
\newblock \bibinfo{journal}{\emph{Advances in neural information processing
  systems}}  \bibinfo{volume}{35} (\bibinfo{year}{2022}),
  \bibinfo{pages}{24824--24837}.
\newblock


\bibitem[Wei et~al\mbox{.}(2023)]%
        {wei2023zero}
\bibfield{author}{\bibinfo{person}{Xiang Wei}, \bibinfo{person}{Xingyu Cui},
  \bibinfo{person}{Ning Cheng}, \bibinfo{person}{Xiaobin Wang},
  \bibinfo{person}{Xin Zhang}, \bibinfo{person}{Shen Huang},
  \bibinfo{person}{Pengjun Xie}, \bibinfo{person}{Jinan Xu},
  \bibinfo{person}{Yufeng Chen}, \bibinfo{person}{Meishan Zhang},
  {et~al\mbox{.}}} \bibinfo{year}{2023}\natexlab{}.
\newblock \showarticletitle{Zero-shot information extraction via chatting with
  chatgpt}.
\newblock \bibinfo{journal}{\emph{arXiv preprint arXiv:2302.10205}}
  (\bibinfo{year}{2023}).
\newblock


\bibitem[Yong et~al\mbox{.}(2023)]%
        {yong2023low}
\bibfield{author}{\bibinfo{person}{Zheng-Xin Yong}, \bibinfo{person}{Cristina
  Menghini}, {and} \bibinfo{person}{Stephen~H Bach}.}
  \bibinfo{year}{2023}\natexlab{}.
\newblock \showarticletitle{Low-resource languages jailbreak gpt-4}.
\newblock \bibinfo{journal}{\emph{arXiv preprint arXiv:2310.02446}}
  (\bibinfo{year}{2023}).
\newblock


\bibitem[Zeng et~al\mbox{.}(2023)]%
        {zeng2023evaluating}
\bibfield{author}{\bibinfo{person}{Zhiyuan Zeng}, \bibinfo{person}{Jiatong Yu},
  \bibinfo{person}{Tianyu Gao}, \bibinfo{person}{Yu Meng},
  \bibinfo{person}{Tanya Goyal}, {and} \bibinfo{person}{Danqi Chen}.}
  \bibinfo{year}{2023}\natexlab{}.
\newblock \showarticletitle{Evaluating large language models at evaluating
  instruction following}.
\newblock \bibinfo{journal}{\emph{arXiv preprint arXiv:2310.07641}}
  (\bibinfo{year}{2023}).
\newblock


\bibitem[Zhao et~al\mbox{.}(2023a)]%
        {CASPER}
\bibfield{author}{\bibinfo{person}{Wei Zhao}, \bibinfo{person}{Zhe Li}, {and}
  \bibinfo{person}{Jun Sun}.} \bibinfo{year}{2023}\natexlab{a}.
\newblock \showarticletitle{Causality analysis for evaluating the security of
  large language models}.
\newblock \bibinfo{journal}{\emph{arXiv preprint arXiv:2312.07876}}
  (\bibinfo{year}{2023}).
\newblock


\bibitem[Zhao et~al\mbox{.}(2023b)]%
        {llm2023survey}
\bibfield{author}{\bibinfo{person}{Wayne~Xin Zhao}, \bibinfo{person}{Kun Zhou},
  \bibinfo{person}{Junyi Li}, \bibinfo{person}{Tianyi Tang},
  \bibinfo{person}{Xiaolei Wang}, \bibinfo{person}{Yupeng Hou},
  \bibinfo{person}{Yingqian Min}, \bibinfo{person}{Beichen Zhang},
  \bibinfo{person}{Junjie Zhang}, \bibinfo{person}{Zican Dong},
  \bibinfo{person}{Yifan Du}, \bibinfo{person}{Chen Yang},
  \bibinfo{person}{Yushuo Chen}, \bibinfo{person}{Zhipeng Chen},
  \bibinfo{person}{Jinhao Jiang}, \bibinfo{person}{Ruiyang Ren},
  \bibinfo{person}{Yifan Li}, \bibinfo{person}{Xinyu Tang},
  \bibinfo{person}{Zikang Liu}, \bibinfo{person}{Peiyu Liu},
  \bibinfo{person}{Jian-Yun Nie}, {and} \bibinfo{person}{Ji-Rong Wen}.}
  \bibinfo{year}{2023}\natexlab{b}.
\newblock \bibinfo{title}{A {{Survey}} of {{Large Language Models}}}.
\newblock
\newblock
\showeprint{2303.18223}~[cs]


\bibitem[Zou et~al\mbox{.}({[n.\,d.]})]%
        {advbench}
\bibfield{author}{\bibinfo{person}{Andy Zou}, \bibinfo{person}{Zifan Wang},
  \bibinfo{person}{Nicholas Carlini}, \bibinfo{person}{Milad Nasr},
  \bibinfo{person}{J.~Zico Kolter}, {and} \bibinfo{person}{Matt Fredrikson}.}
  \bibinfo{year}{[n.\,d.]}\natexlab{}.
\newblock \bibinfo{booktitle}{\emph{Universal and Transferable Adversarial
  Attacks on Aligned Language Models}}.
\newblock
\showeprint[arxiv]{2307.15043}~[cs]
\urldef\tempurl%
\url{http://arxiv.org/abs/2307.15043}
\showURL{%
\tempurl}


\end{thebibliography}
\end{document}